\documentclass[lettersize,journal]{IEEEtran}
\usepackage{amsmath,amsfonts}
\usepackage[ruled,lined]{algorithm2e}
\usepackage{array}
\usepackage{xcolor}
\usepackage[caption=false,font=normalsize,labelfont=sf,textfont=sf]{subfig}
\usepackage{textcomp}
\usepackage{stfloats}
\usepackage{url}
\usepackage{hyperref}
\usepackage{verbatim}
\usepackage{booktabs}
\usepackage{graphicx}
\usepackage{tabularx}
\usepackage{cite}
\usepackage{multirow}
\DeclareMathOperator*{\argmax}{arg\,max}

\begin{document}

\title{Multi-label Scene Classification for Autonomous Vehicles: Acquiring and Accumulating Knowledge from Diverse Datasets}
%\title{Learning Multi-label Scene Classification for Autonomous Vehicles from Diverse Datasets}

\author{Ke Li, ~\IEEEmembership{Graduate Student Member,~IEEE},
Chenyu Zhang, ~\IEEEmembership{Graduate Student Member,~IEEE},
Yuxin Ding, 
Xianbiao Hu, 
Ruwen Qin\IEEEauthorrefmark{1},~\IEEEmembership{Member,~IEEE}% <-this % stops a space
\thanks{Ke Li, Chenyu Zhang, and Ruwen Qin are with the Department of Civil Engineering, Stony Brook Univerisity, Stony Brook, NY 11794, USA.}% <-this % stops a space
\thanks{Yuxin Ding and Xianbiao Hu are with the Department of Civil and Environmental Engineering, Pennylvania State University, University Park, PA 16802, USA.}
\thanks{\IEEEauthorrefmark{1} Corresponding author: Ruwen Qin, email: ruwen.qin@stonybrook.edu}
\thanks{Manuscript received Month dd, yyyy; revised Month dd, yyyy.}
}

% The paper headers
\markboth{Journal of ,~Vol.~xx, No.~x, Month~yyyy}{Shell \MakeLowercase{\textit{Li et al.}}: A Sample Article Using IEEEtran.cls for IEEE Journals}

%\IEEEpubid{0000--0000/00\$00.00~\copyright~2021 IEEE}
% Remember, if you use this you must call \IEEEpubidadjcol in the second
% column for its text to clear the IEEEpubid mark.

\maketitle

\begin{abstract}
Driving scenes are inherently heterogeneous and dynamic. Multi-attribute scene identification, as a high-level visual perception capability, provides autonomous vehicles (AVs) with essential contextual awareness to understand, reason through, and interact with complex driving environments. Although
scene identification is best modeled as a multi-label classification problem via multitask learning, it faces two major challenges: the difficulty of acquiring balanced, comprehensively annotated datasets and the need to re-annotate all training data when new attributes emerge. To address these challenges, this paper introduces a novel deep learning method that integrates Knowledge Acquisition and Accumulation (KAA) with Consistency-based Active Learning (CAL). KAA leverages monotask learning on heterogeneous single-label datasets to build a knowledge foundation, while CAL bridges the gap between single- and multi-label data, adapting the foundation model for multi-label scene classification. An ablation study on the newly developed Driving Scene Identification (DSI) dataset demonstrates a 56.1\% improvement over an ImageNet-pretrained baseline. Moreover, KAA-CAL outperforms state-of-the-art multi-label classification methods on the BDD100K and HSD datasets, achieving this with 85\% less data and even recognizing attributes unseen during foundation model training. The DSI dataset and KAA-CAL implementation code are publicly available at \url{https://github.com/KELISBU/KAA-CAL}.

%The capability to identify driving scenes using various attributes is a high-level perception crucial for autonomous vehicles (AVs), as it provides the contextual awareness necessary for AVs to understand, reason through, and interact with their complex driving environment. 
\end{abstract}


\section{Introduction}

\begin{figure*}[htbp]
    \centering
    \includegraphics[width=\linewidth]{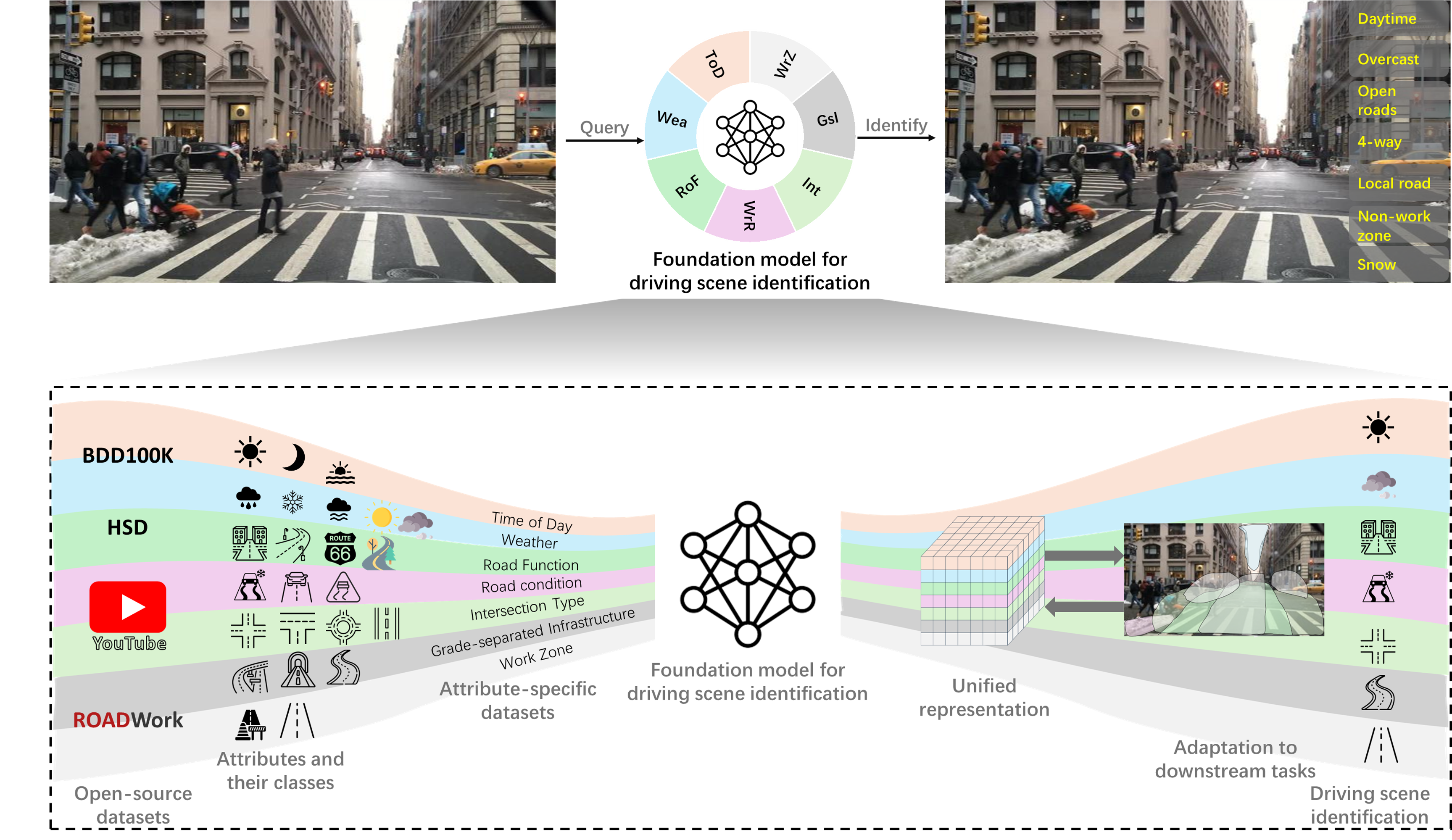}
    \caption{A foundation model for multi-attribute driving scene identification, with its comprehensive knowledge obtained from diverse single-label datasets}
    \label{fig:FoundationModelConcept}
\end{figure*}

\IEEEPARstart{D}{riving} scene identification involves assigning non-exclusive labels to each scene captured by an onboard camera. As illustrated in Fig. \ref{fig:FoundationModelConcept},  a complex scene is characterized using various attributes, such as time of day, weather, roadway function, intersection type, among others. Scene identification is a high-level perception crucial for autonomous vehicles (AVs), as it provides the contextual awareness for reasoning and decision-making \cite{IV1,IV2}. For example, the friction coefficient of the road surface varies depending on whether the road is dry, wet, snowy, or icy. Identifying driving scenes based on the road surface condition allows vehicles to adjust their estimations of braking distance and turning speed accordingly, thus improving safety. Driving scene identification also supports data generation, as it provides scene attribute labels that serve as semantic priors to guide image and video synthesis. For example, after aligning visual and semantic representations, the intersection type class serves as an explicit conditioning information for fine-grained generation (e.g., three-way, four-way, or roundabout).

Driving scenes are inherently heterogeneous and dynamic, requiring the simultaneous recognition of many attributes. This makes scene identification a multi-label classification problem that goes far beyond a simple perception task. The complexity nature of this important problem highlights the critical need for a foundation model possessing the comprehensive knowledge to accurately characterize these intricate scenes. 

While seemingly straightforward for driving scene identification, standard multitask learning methods face critical limitations. A primary challenge is the lack of a comprehensively annotated multi-label training dataset. In a high-dimensional attribute space, driving scene distribution is highly imbalanced \cite{multi2, multi3}. Some scenarios are rare, making it difficult to collect balanced multi-label training samples. This likely explains why image- and video-level annotations in popular driving scene datasets are either single-labeled (e.g., \cite{Wu2019WZTraffic, workzone}) or multi-labeled with just limited scene attributes (e.g., \cite{yu2020bdd100k, narayanan2019dynamic}), and why existing models are predominantly single-label classification models (e.g., \cite{single1,single2,scene1}). Meanwhile, as the transportation system evolves, the need to identify additional attributes constantly emerges. However, the standard multitask learning framework proves inefficient for integrating desired new knowledge. This is because it requires not only re-labeling entire training datasets with the required new labels but also a further data collection, a process that is neither scalable nor cost-effective.

The aforementioned challenges, combined with the abundance of single-label datasets and the convenience of monotask learning, motivate the exploration of a learning approach leading to a driving scene foundation model. As illustrated in Fig. \ref{fig:FoundationModelConcept}, this strategy constructs the model by learning from heterogeneous datasets and consolidating transferable representations. The resulting model can not only recognize complex driving scenes using a variety of attributes but also quickly learn to recognize new ones. 

The learning process to be introduced in this paper is inspired by how students learn. A student model concentrates on learning one classification task from a corresponding single-label dataset at a time in a sequential manner and consolidates the acquired knowledge in a teacher model, which serves as a reference for the student model's continual learning and improvement. This sequential process goes cyclically, progressively transforming knowledge from heterogeneous single-label datasets into a unified representation, forming a foundation model for multi-attribute driving scene identification. Yet, this approach confronts an inevitable domain shift problem, which arises from the discrepancy between the marginal distributions of individual attributes and their joint distribution. To mitigate this, a strategy is developed to guide the model's adaptation, transforming the knowledge it acquires from diverse single-label datasets into a comprehensive understanding suitable for complex scene recognition.

This paper reports an effort in designing and validating the described new deep learning method for complex driving scene identification. In addressing challenges confronted in this study, the paper makes the following contributions:
\begin{itemize}
    \item A new deep learning system, \textbf{K}nowledge \textbf{A}cquisition and \textbf{A}ccumulation (KAA), is proposed to learn from heterogeneous single-label datasets and consolidate learned knowledge into a unified representation, laying the groundwork for multi-label driving scene identification.
    \item In demonstrating KAA, a comprehensive and scalable, \textbf{D}riving \textbf{S}cene \textbf{I}dentification (DSI) dataset is constructed, comprising seven single-label subsets, each annotated according to one specific scene attribute, representing  heterogeneous sources of information about driving scenes.     
    \item {An adaptation algorithm, \textbf{C}onsistency-based \textbf{A}ctive \textbf{L}earning (CAL), is developed to couple with KAA, addressing the domain shift problem it confronts.}
\end{itemize}

Remainder of the paper is organized as follows. Section \ref{related} summarizes recent work related to this study. Then, Section \ref{sec:datasets} details the components of the new DSI dataset. After that, the proposed KAA-CAL learning method is introduced in Section \ref{sec:Methodology}, with the implementation delineated in Section \ref{sec:Implementation Details}. Section \ref{resultandexper} further reports the experimentation and results. At the end, Section \ref{conclusion} concludes the study by summarizing important findings and future research directions. 

\IEEEpubidadjcol

\section{Related Work}\label{related}

This paper is built upon a spectrum of research areas. The most relevant literature from these fields is summarized below.

\subsection{Public Driving Scene Datasets}

Various driving video datasets have been developed to support different perception tasks for AVs, including scene classification. BDD100K \cite{yu2020bdd100k} is a large-scale driving video dataset featuring image-level annotations for six weather conditions, six scene types, and three distinct times of day. Honda Scene Dataset (HSD) \cite{narayanan2019dynamic} contains 80 hours of driving video data clips collected in the San Francisco Bay Area. HSD provides video-level identification of eleven road places, four categories of road environment, and four weather conditions, thereby broadening the diversity of both attributes and classes for urban scene identification. ROADWork \cite{workzone} is a public dataset focused on work zones identification with data collected from eighteen cities in the U.S. This dataset also provides fine-grained annotations for instance and semantic segmentation. DENSE++ \cite{RECNET} provides multi-label annotations for environmental conditions including daytime, precipitation, fog, road condition, roadside condition, and scene-setting for 12,997 images collected from Northern Europe. While the annotation method is provided, the image labels are not publicly available. WZ-Traffic \cite{Wu2019WZTraffic} is a scene dataset with 6,035 single-label images in 20 categories. Other open-source driving datasets, including Cityscapes \cite{Cordts2016Cityscapes}, KITTI \cite{KITTI}, and nuScenes \cite{nuscenes2019}, are primarily annotated for object detection, semantic segmentation, and instance segmentation tasks. 

Few datasets provide comprehensive annotations for multi-label scene classification. Some provide class labels either at the frame level or the video level, yet they do so based on one or a few attributes of their interest. Integrating the information in heterogeneous datasets as a unified representation of driving scenes is greatly desired.   
%Additionally, there is no common standard or widely accepted taxonomy for categorizing driving scenes with respect to any given scene attribute, leading to inconsistencies in scene classes across different datasets.

\subsection{Multi-label Driving Scene Classification}
\label{subsec:Multi-label Driving Scene Classification}

Multi-attribute scene identification is commonly modeled as a multi-label image classification proble,. A few studies have tackled this problem in various approaches. Duong et al. \cite{multi2} developed CF-Net, which uses single-label classifiers to enhance the main multi-label classifier by utilizing feature fusion and stacking. The model was trained on a modified BDD100K dataset with three attributes: location, weather, and time of day. Chen et al. \cite{multi3} proposed to solve the multi-label scene classification problem by incorporating the single-label training procedure into the multi-label architecture. Additionally, a deep data integration strategy was utilized to improve the classification ability. RECNet \cite{RECNET} proposed a hierarchical strategy for annotating six environmental conditions, including daytime, fog, precipitation, road condition, roadside condition, and infrastructure, leading to a fully annotated dataset DENSE++. This dataset was used to train a multi-label classification model, which uses the EfficientNet-B2 as the shared backbone for the six downstream classifiers. Prykhodchenko and Skruch \cite{MUlti5} also trained a multi-label scene classification model on BDD100K via multitask learning. Data augmentation and balancing strategies were utilized to address the class imbalance issue in the BDD100K dataset.
 
Existing methods generally adhere to the standard multitask learning framework, thus relying on multi-label training datasets. New methodologies that can transcend this framework will offer an opportunity to utilize the abundant driving scene images or video data.

\subsection{Knowledge Distillation}

Knowledge distillation (KD) typically utilizes a teacher-student framework to distill the knowledge accumulated in a deep or large model (the teacher) into a smaller or shallow one (the student). It supports the student network's continual acquisition and consolidation of scene identification knowledge.

KD methods are generally categorized into three types: response-based KD, where the student directly mimics the teacher's final predictions; feature-based KD, which transfers knowledge by leveraging features extracted by the teacher; and relation-based KD, which explores relationships between different layers or samples \cite{knowledge1}. When both the teacher and student networks are deep, regulating multiple hidden layers of the student network leads to improved performance compared to response-based KD \cite{romero2014fitnets}. Ma et al. \cite{ark} introduced ARK to accrue and reuse feature-based knowledge, addressing the challenge of inconsistent annotations in various medical image datasets. Using ARK, a foundation model is constructed for medical image analysis. When KD occurs progressively, there is a risk of forgetting previously acquired knowledge. To mitigate this issue, Van de Ven et al. \cite{brain} proposed a brain-inspired method that replays internal or hidden representations in continual learning. While sharing some similarities with proposed KAA, ARK primarily focuses on addressing annotation heterogeneity across different medical datasets. 

\subsection{Deep Active Learning}\label{deepactive}

Deep active learning can efficiently adapt models to new domains or tasks, achieved by strategically selecting the most informative and representative samples for adaptation. Some uncertainty-based methods such as maximum entropy, margin sampling, least-confidence sampling, and Bayesian Active Learning by Disagreement rely on uncertainty measures based on a task \cite{DALsurvey}. In contrast, a task-agnostic method introduced in \cite{learningloss} integrates a loss prediction module to select top samples of the highest prediction loss from the unlabeled pool. Uncertainty-based methods may cause sample redundancy. 

A diversity-based method selects instances from the unlabeled pool to represent a broad data distribution. Beyond traditional clustering methods like K-means, Core-Set Section \cite{coreset} computes the distance of features from a designated hint layer within a deep learning model. A greedy search algorithm is employed to iteratively select samples that exhibit the greatest distance from their nearest neighbors until the selection budget is exhausted. Diversity-based methods may overlook the most informative instances.  

Hybrid methods, including Wasserstein Adversarial Active Learning and Batch Active learning by Diverse Gradient Embeddings, effectively address limitations of both uncertainty- and diversity-based methods \cite{DALsurvey}. For example, Hekimoglu et al. \cite{Hekimoglu_2024_WACV} quantified uncertainty using an inconsistency score computed as the maximum loss between the initial-refined pairwise task predictions. Simultaneously, a reconstructed feature embedding that condenses information across all tasks measures the diversity score. The combination of these two components forms the basis for deep active learning. Yet this approach cannot handle the domain shift caused by monotask learning from diverse single-label datasets.

\section{Driving Scene Identification (DSI) Dataset}
\label{sec:datasets}

The data distribution in the high-dimensional scene attribute space is highly imbalanced. Some combinations of scene classes are corner cases difficult to collect data for. This limits the creation of comprehensively annotated datasets to directly train multi-label scene classification models via multitask learning. However, collecting data from respective marginal distributions of scene attributes is relatively straightforward. 

\begin{figure*}[!t]
\centering
\includegraphics[width=2\columnwidth]{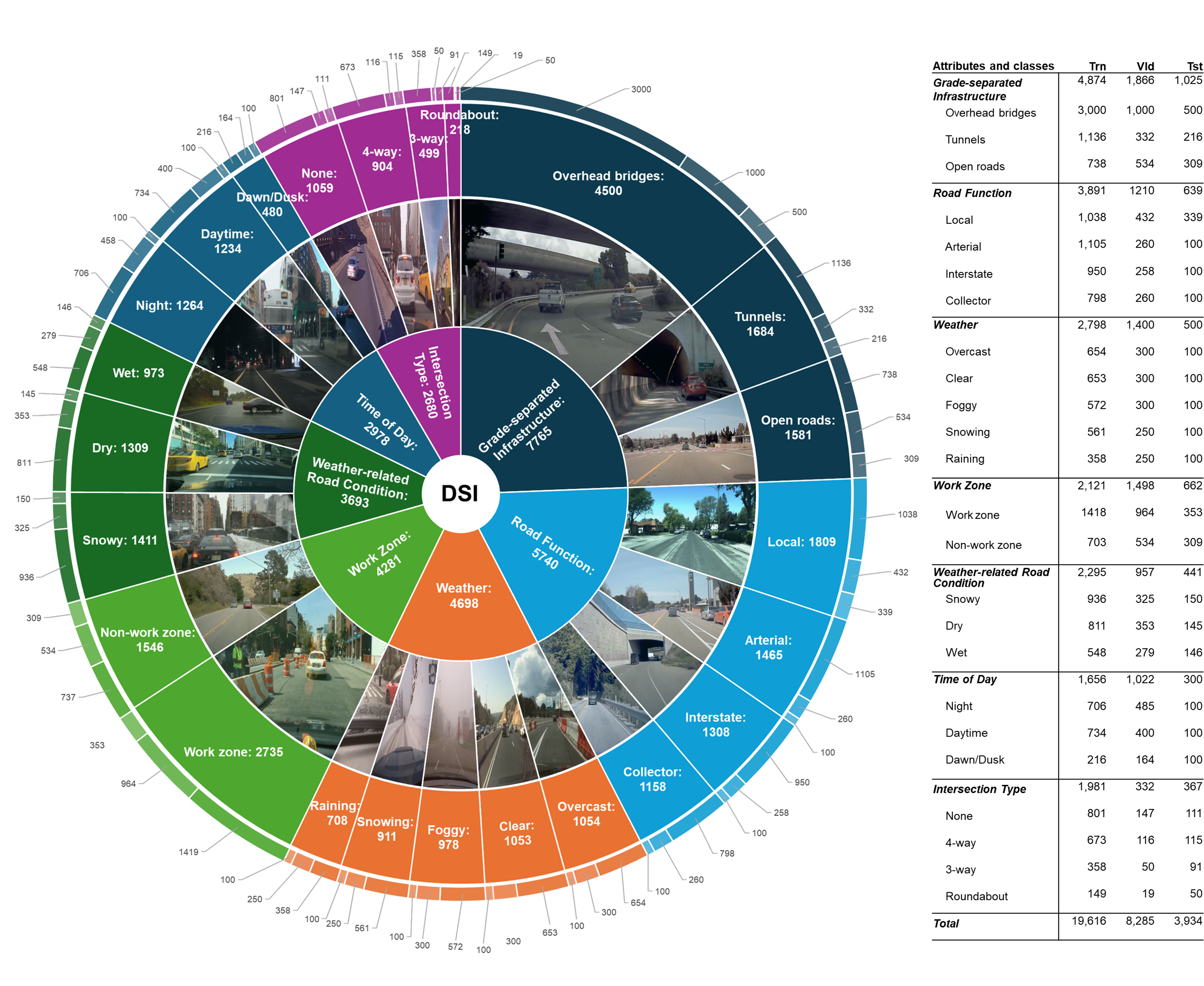}
\caption{Overview of the DSI dataset}
\label{fig:dataset}
\end{figure*}

This paper introduces DSI, a heterogeneous single-label dataset for demonstrating  KAA-CAL, the proposed learning method. DSI consists of 31,835 scene images sampled from public driving video datasets, including BDD100K \cite{yu2020bdd100k}, HSD \cite{narayanan2019dynamic}, and ROADWork Data \cite{workzone}, and is supplemented with additional images sampled from YouTube videos. An overview of the DSI dataset and its statistics are presented in Fig. \ref{fig:dataset}. The dataset is available to the public at \url{https://github.com/KELISBU/KAA-CAL}.

%\subsection{Driving Scene Attributes and Classes}

The DSI dataset consists of $M$ (7) subsets, denoted as $\{\mathcal{D}_m\}_{m=1}^M$, where $m$ is the index of attributes for scene identification. Each subset is annotated according to one unique scene attribute and further partitioned into training (Trn), validation (Vld), and test (Tst) sets:
\begin{equation}
    \mathcal{D}_m = \mathcal{D}_m^{\text{Trn}}\cup\mathcal{D}_m^{\text{Vld}}\cup\mathcal{D}_m^{\text{Tst}}.
    \label{eq:D_m}
\end{equation}

Fig. \ref{fig:dataset} illustrates the scene attributes and respective classes. In total, DSI provides image-level labels for 24 classes, with each falling exclusively under one of the seven scene attributes. Classifying driving scenes according to these scene attributes offers valuable insights for AVs. Notably, DSI can be conveniently expanded by adding new single-label subsets or new classes within existing subsets (e.g., a new design of intersection type) as needs for new knowledge about scene identification emerge.

%As illustrated in Fig. \ref{fig:dataset}, these scene attributes include Grade-separated Infrastructure, Road Function, Weather, Work Zone, Weather-related Road Condition, Time of Day, and Intersection Type. In total, DSI provides image-level labels for 24 classes, with each falling exclusively under one of the seven scene attributes. Classifying driving scenes according to these scene attributes offers valuable insights for AVs.

{\bf Grade-separated Infrastructure}. Road segments featuring grade-separated infrastructure, such as bridges and tunnels, often present specific challenges for AVs. Bridges above the road may limit the sensor field of view, cause sudden lighting changes, and introduce gusts of wind. Tunnels, on the other hand, have different lighting conditions, limited visibility, and potential congestion. The Grade-separated Infrastructure subset in DSI comprises three classes: Overhead bridges, Tunnels, and Open roads, totaling 7,765 images. This subset was curated from the HSD dataset using query keywords like ``overhead bridge'' and ``tunnel''. To ensure these structures are visually perceivable, additional query criteria such as ``Approaching'' and ``Entering'' were used for data sampling.

{\bf Road Function}. Each road type has specific characteristics, such as a speed limit range, traffic density, vehicle type distribution, terrain, and travel purposes. Thus, understanding the road function assists adaptive driving behavior. The US road functional system has not been systematically annotated in public datasets. To address this, the Road Function subset was created by querying with keywords like ``road function'' on YouTube and sampling every fifth frame from the collected videos. This dataset consists of 5,740 images across 4 classes: Local, Arterial, Interstate, and Collector. 

{\bf Weather}. Identification of the weather condition is critical because an adverse weather condition risks driving safety and the performance of visual tasks. The Weather subset includes 4,698 frames sampled from the BDD100K dataset, encompassing Overcast, Clear, Foggy, Snowing, and Raining conditions. To mitigate the issue of data imbalance caused by the rarity of foggy scenes, synthetic images were generated to ensure a more balanced distribution across weather conditions.

{\bf Work Zone}. Driving safety is a concern in work zones, where construction equipment, temporary barriers, road workers, and other related facilities are present. Work zones often involve lane closures. Confusion arises when old, not fully removed lane markings mix with new ones, especially for AVs. Identifying work zones allows AVs to adjust their lane position and speed dynamically, minimizing crash risks. The training and validation data of the Work Zone subset mainly come from BDD100K and HSD, and its testing data are primarily from the ROADWork dataset. 

{\bf Weather-related Road Condition}. The friction coefficient of the road surface changes depending on weather conditions like snow or rain. Identifying these weather-related road conditions allows AVs to adjust their control for better adaptation to the specific surface characteristics. The Weather-related Road Condition subset includes 3,693 images collected from both YouTube and BDD100K, intended for identifying Snowy, Dry, and Wet conditions.

{\bf Time of Day}. The lighting condition directly impacts the visibility of objects, roads, and the surrounding environment because illumination influences key visual functions like contrast sensitivity, visual acuity, depth perception, and peripheral vision. Consequently, the lighting condition affects not only human drivers' visual perception and reaction times but also machine vision due to the affected quality of camera-captured data. For example, nighttime and dawn/dusk images often exhibit increased noise, reduced contrast, and color distortion. The Time of Day subset comprises 2,978 images labeled as Night, Daytime, or Dawn/Dusk, sampled from the BDD100K dataset. 

{\bf Intersection Type}. More than half of the combined total of fatal and injury crashes occur at or near intersections due to increased conflict points and complex interactions among vehicles, pedestrians, and other road users. Recognizing the type of intersection allows AVs to anticipate potential conflict points and make informed decisions, such as yielding, stopping, or merging in complex environments. The Intersection Type subset consists of 2,680 images, categorized into four classes: None, 3-way, 4-way, and Roundabout. The data were collected by integrating YouTube videos, which provide roundabout images, along with the HSD and BDD100K datasets, which contribute images of none, 3-way, and 4-way intersections. 3-way and 4-way intersections are typically more noticeable when ego-vehicles are entering the intersection. Therefore, the query keyword ``entering'' was used when searching for such images in the HSD dataset.

%\subsection{Diversity and Generalization}

%Small biases in training data can lead to poor generalization of deep learning models, as a biased dataset may cause the model to learn features that deviate from the intended task \cite{NEURIPSdata}. Therefore, the data collection heavily emphasized diversity and generalization to mitigate potential biases. For instance, in the Work Zone subset, data points were sampled across various land use categories (urban and rural areas), road functions, and lighting conditions. This ensures that the DSI dataset provides a good opportunity for learning representations likely encountered in the real world.

\section{Methodology}
\label{sec:Methodology}

Multi-label driving scene classification involves assigning multiple non-exclusive class labels to each input scene image $\pmb{x}$ in the domain 
$\mathcal{I}=\{\pmb{x}|\pmb{x}\in\mathbb{R}^{3\times H\times W}\}$, comprising color images with three color channels, a height of $H$ pixels, and a width of $W$ pixels. $\pmb{x}$ can be classified based on $M$ different scene attributes, $Y_m$, for $m=1,\dots,M$. Each attribute $Y_m$ is a categorical variable with its support $\mathcal{U}_m$ comprising the exclusive classes for this attribute. A multi-label classification model $\mathcal{C}$ approximates the mapping, $\mathcal{I}\rightarrow \mathcal{U}_1\times \dots \times \mathcal{U}_M$, which estimates the class labels for each scene image $\pmb{x}$:  
\begin{equation}
[\widehat{\pmb{y}}_1,\dots,\widehat{\pmb{y}}_M]=\mathcal{C}(\pmb{x}; \pmb{\Theta}),
\label{eq:classification model}
\end{equation}
where $\widehat{\pmb{y}}_m \in  \mathbb{R}^{|\mathcal{U}_m|}$ is the prediction in probability for the one-hot encoded true label $\pmb{y}_m\in \mathbb{B}^{|\mathcal{U}_m|}$ of $\pmb{x}$, and $\pmb{\Theta}$ represents the learnable parameters of the classification model $\mathcal{C}$. To build such a model, this section proposes a new deep learning method named KAA-CAL. First, KAA learns a model $\widetilde{\mathcal{C}}=\mathcal{C}(\cdot,\widetilde{\pmb{\Theta}})$ that maximizes the classification accuracy in isolated per-attribute identification. Following this, CAL further adapts $\widetilde{\mathcal{C}}$ to become $\mathcal{C}^\ast=\mathcal{C}(\cdot,\pmb{\Theta}^{\ast})$, one that is proficient in simultaneous all-attribute identification.

\begin{figure*}[t]
\centering
\includegraphics[width=6 in]{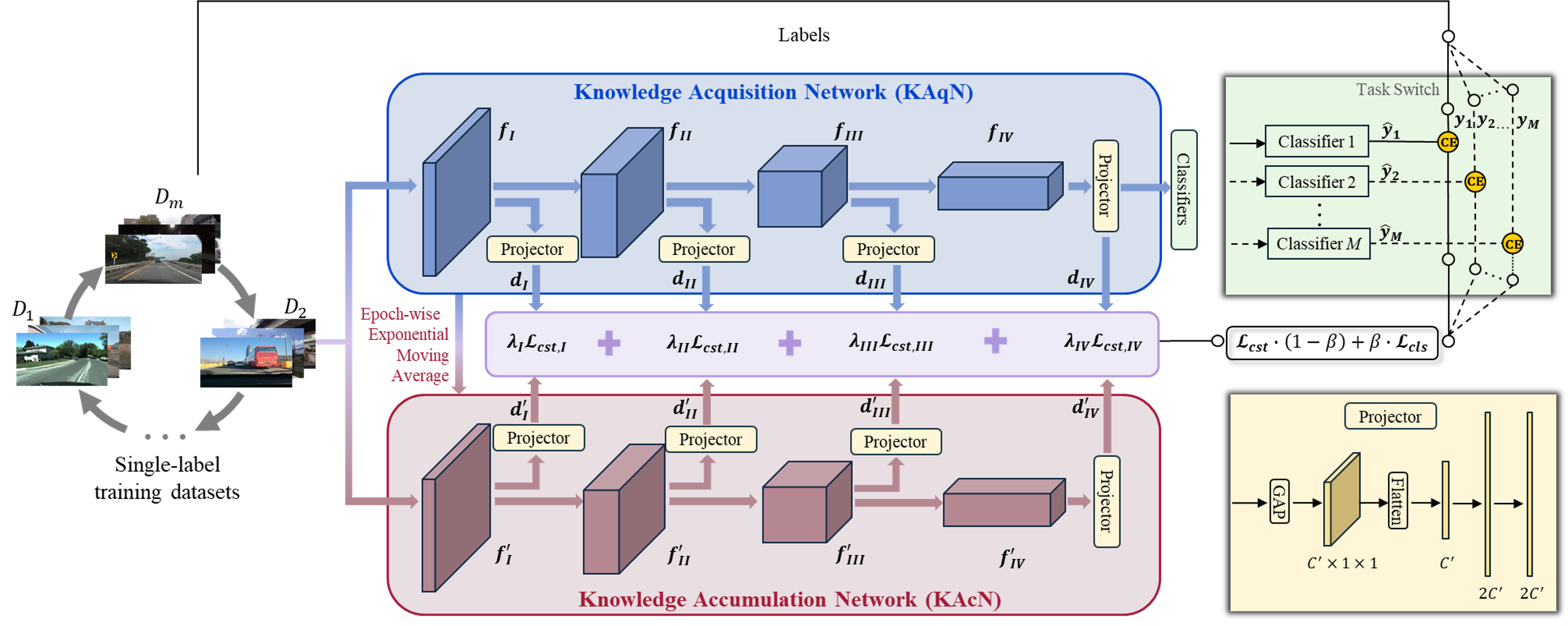}
\caption{Overview of the proposed Knowledge Acquisition and Accumulation (KAA)} learning system \label{fig:methodology}
\end{figure*}

\subsection{Knowledge Acquisition and Accumulation (KAA)}\label{subsec:KAA}

\subsubsection{Overview of the Learning Process}
KAA, depicted in Fig. \ref{fig:methodology}, is a deep learning system designed to acquire and accumulate knowledge from various single-label datasets for multi-label driving scene classification. KAA utilizes a teacher-student network architecture. The knowledge acquisition network (KAqN) serves as the student network, responsible for learning new knowledge and consolidating it into the existing knowledge in the learning system. The knowledge accumulation network (KAcN), acting as the teacher network, stores the consolidated knowledge and applies it to guide the student network in continual learning and improvement.
%\cite{eichenbaum2017prefrontal}

KAqN's deep encoder is trained to attain the knowledge for performing $M$ downstream tasks, indexed by $m$, each focused on classifying the input driving scene image according to one attribute. The overall approach of KAqN is a sequential and cyclical learning process, as outlined in Algorithm \ref{alg:KAAN}. A training epoch, indexed by $t$, is one learning cycle. During each learning cycle, KAqN goes through $M$ learning iterations to sequentially learn on these classification tasks, utilizing a dedicated single-label dataset for each task. Therefore, the learning iteration $i$ corresponds to a specific learning cycle $t$ and learning task $m$:
\begin{equation}
\begin{aligned}
    &t = \lceil i/M \rceil,\\
    &m = i - (t-1)M.
\end{aligned} 
\end{equation}

Upon completion of one learning cycle, the newly acquired knowledge is consolidated into the existing knowledge stored in KAcN, forming an updated foundation for continual learning in the subsequent cycle. This cyclical and sequential learning process continues until the stopping criterion is met. KAA's architecture and design are further delineated below. 

\begin{algorithm}[t]
\caption{KAA Algorithm}\label{alg:KAAN}
\SetKwInput{In}{INPUT}
    \SetKwInput{Out}{OUTPUT}
\SetKwInput{Init}{INITIALIZATION}
\SetKwInput{Cycle}{CYCLICAL \& SEQUENTIAL LEARNING PROCESS}

 \In{\\
 $\{\mathcal{D}_m \mid m=1,\dots,M\}$:the DSI dataset comprising $M$ single‐label subsets}
  
  \BlankLine 

\Init{\\ 
    $i \leftarrow 0$\text{: index of learning iteration}\\
    $t \leftarrow 0$: index of learning cycle\\
    $\pmb{\theta}(0)$: initial values for KAqN\\
    $\pmb{\phi}_m(0)$: initial values for classifier $m$, for $m=1,\dots,M$\\
    $\pmb{\theta}'(0)$: initial values for KAcN}
\BlankLine 

\Cycle{\\
$i=0$\\
\For{cycle $t=1,\dots,T$}{
\BlankLine 
    {KAqN acquires new knowledge sequentially:}\\
    \For{task $m=1,\dots,M$}{
        $i = i + 1$\\
        learning classification task $m$ on dataset $\mathcal{D}_m$:\\
        \Indp
        $\pmb{\theta}(i-1)$ is updated to become $\pmb{\theta}(i)$\\
        $\pmb{\phi}_m(t-1)$ is updated to become $\pmb{\phi}_m(t)$\\
    \Indm}
    \BlankLine 
    {$\pmb{\theta}(Mt)$ is consolidated with $\pmb{\theta}^{'}(t-1)$:}\\
    \Indp
    $\pmb{\theta}'(t)\xleftarrow{\pmb{\theta}(Mt)}\pmb{\theta}^{'}(t-1)$\\
    \Indm
    \BlankLine 
    {Termination of the learning process:}\\
    \If{stopping criterion is satisfied}{
       $\tilde t = t$\;
       \textbf{break}\\
    }
}}
\Out{$\widetilde{\pmb{\Theta}}\leftarrow\{\pmb{\theta}^{'}(\tilde{t}), \pmb{\phi}_1(\Tilde{t}),\dots, \pmb{\phi}_M(\tilde{t})\}$}
\end{algorithm}

\subsubsection{Knowledge Acquisition}

KAqN utilizes the backbone of Swin-B \cite{liu2021swin}, a vision transformer pretrained on ImageNet, as the deep feature encoder. The encoder embeds each input image as a feature map for the downstream classification tasks. 
Specifically, Swin-B processes an input scene image with a resolution of $244\times244$ by first partitioning it into non-overlapping $4\times4$ patches. It then sequentially encodes these patches into a feature map through four stages, progressively reducing the spatial resolution of the
input scene image while increasing the depth (number of channels) of the feature maps. Each stage applies either a linear embedding ($\operatorname{LE}$) or a patch merging ($\operatorname{PM}$) layer, followed by multiple Swin Transformer ($\operatorname{ST}$) blocks:
\begin{equation}
    \begin{aligned}
    &\pmb{f}_{\textsc{i}}=\operatorname{ST}(\operatorname{ST}(\operatorname{LN}(\pmb{x}; \pmb{\theta}^{\mathcal{E}}_{\text{I},0});\pmb{\theta}^{\mathcal{E}}_{\textsc{i},1});\pmb{\theta}^{\mathcal{E}}_{\textsc{i},2}),\\
    &\pmb{f}_{\textsc{ii}}=\operatorname{ST}(\operatorname{ST}(\operatorname{PM}(\pmb{f}_{\textsc{i}}; \pmb{\theta}^{\mathcal{E}}_{\text{II},0});\pmb{\theta}^{\mathcal{E}}_{\textsc{ii},1});\pmb{\theta}^{\mathcal{E}}_{\textsc{ii},2}),\\
    &\pmb{f}_{\textsc{iii}}=\operatorname{ST}(\dots\operatorname{ST}(\operatorname{PM}(\pmb{f}_{\textsc{ii}}; \pmb{\theta}^{\mathcal{E}}_{\text{III},0});\pmb{\theta}^{\mathcal{E}}_{\textsc{iii},1});\pmb{\theta}^{\mathcal{E}}_{\textsc{iii},6}),\\
    &\pmb{f}_\textsc{iv}=\operatorname{ST}(\operatorname{ST}(\operatorname{PM}(\pmb{f}_{\textsc{iii}} ;\pmb{\theta}^{\mathcal{E}}_{\text{IV},0});\pmb{\theta}^{\mathcal{E}}_{\textsc{iv},1});\pmb{\theta}^{\mathcal{E}}_{\textsc{iv},2}).
    \end{aligned}
\label{eq:Swin-B}
\end{equation}
Here, 
$\pmb{\theta}^{\mathcal{E}}$ designates all the learnable parameters of KAqN's deep encoder. The extracted feature maps have the following dimensions: $\pmb{f}_\textsc{i} \in \mathbb{R}^{128\times 56 \times 56}$, $\pmb{f}_{\textsc{ii}} \in \mathbb{R}^{256\times 28 \times 28}$, $\pmb{f}_{\textsc{iii}} \in \mathbb{R}^{512\times 14 \times 14}$, and $\pmb{f}_{\textsc{iv}} \in \mathbb{R}^{1024 \times 7 \times 7}$. 

%The $M$ classifiers for KAqN have an identical structure. 

Then, four projectors respectively embed the individual feature maps derived from each stage as feature vectors, 
\begin{equation}
    \pmb{d}_s =  \operatorname{MLP}(\operatorname{Flatten}(\operatorname{GAP}(\pmb{f}_s)); \pmb{\theta}^{\mathcal{P}}_{s}),
    \label{eq:projector}
\end{equation}
for $s = \textsc{I}, \dots, \textsc{IV}$. Here, $\pmb{\theta}^\mathcal{P}$ designates the learnable parameters for the four projectors. Eq. (\ref{eq:projector}) states that each projector first downsamples the input feature map $\pmb{f}_s$ through global average pooling ($\operatorname{GAP}$), reducing its spatial dimensions to 1 while keeping the number of channels unchanged. Then, the downsampled feature map is flattened ($\operatorname{Flatten}$) as a vector, which then goes through a multi-layer perceptron ($\operatorname{MLP}$) with two hidden layers. The resulting feature embeddings are $\pmb{d}_\textsc{i} \in \mathbb{R}^{256}$, $\pmb{d}_{\textsc{ii}} \in \mathbb{R}^{512}$, $\pmb{d}_{\textsc{iii}} \in \mathbb{R}^{1024}$, and $\pmb{d}_{\textsc{iv}} \in \mathbb{R}^{2048}$.

%The first hidden layer has 1,024 neurons and the second layer has 2,048 neurons, and both use $\operatorname{ReLU}$ as the activation function. 

Finally, the feature embedding from the last stage, $\pmb{d}_\textsc{iv}$, flows into classifier $m$, which is active in the current learning iteration, to predict the probability distribution for the classes of attribute $m$:
\begin{equation}
    \widehat{\pmb{y}}_m= \operatorname{SM}(\operatorname{FC}(\pmb{d}_{\textsc{iv}};\pmb{\phi}_{\operatorname{FC},m});\pmb{\phi}_{\operatorname{SM},m}).
    \label{eq:classifier}
\end{equation}
The fully-connected (FC) layer of the classifier reduces the dimension of feature embedding $\pmb{d}_{\textsc{iv}}$, and the softmax operator ($\operatorname{SM}$) normalizes the output as the probability distribution over classes. $\pmb{\phi}_{m}$ in Eq. (\ref{eq:classifier}) represents the learnable parameters of classifier $m$.

%Eq. (\ref{eq:classifier}) The extracted task-specific feature vector, after passing through a fully connected layer, $\operatorname{FC}(\cdot)$, and a softmax operator, $\operatorname{SM}$, becomes the probability distribution of the classes of attribute $m$,  $\widehat{\pmb{y}}_m$. 
% where $\pmb{\theta}_{m}$ represents the learnable parameters of classifier $m$. Eq. (\ref{eq:classifier}) shows the classifier first downsamples $\pmb{f}_\textsc{iv}$ through global average pooling, $\operatorname{GAP}(\cdot)$, which reduces its spatial dimensions from $7\times 7$ to 1, while keeping the number of channels unchanged. Then, the downsampled feature vector goes through a multilayer perceptron with two hidden layers, $\operatorname{MLP}(\cdot)$, which extracts features for classifying the scene based on attribute $m$. The extracted task-specific feature vector, after passing through a fully connected layer, $\operatorname{FC}(\cdot)$, and a softmax operator, $\operatorname{SM}$, becomes the probability distribution of the classes of attribute $m$,  $\widehat{\pmb{y}}_m$. 

KAqN's performance on the classification task $m$ in any learning iteration is evaluated by calculating the corresponding cross-entropy loss:
\begin{equation}
    \mathcal{L}_{cls,m}  = - \pmb{y}_m \cdot \log\widehat{\pmb{y}}_m,
    \label{eq:cross-entropy task}
\end{equation}
where $\cdot$ denotes the inner product of the one-hot encoded ground truth class label, $\pmb{y}_m$, and the predicted probability distribution over classes, $\widehat{\pmb{y}}_m $. The prediction loss in Eq. (\ref{eq:cross-entropy task}) is part of the total loss for guiding the modeling training. 

%Minimizing the regularized version of this prediction loss (see Eq. (\ref{eq:total_loss})) with respect to KAqN's learnable parameters $\pmb{\theta}$ (comprising both $\pmb{\theta}^{\mathcal{E}}$ and $\pmb{\theta}^{\mathcal{P}}$) and the classifier's learnable parameters $\pmb{\phi}_m$ in a learning iteration will improve its ability to perform this task.

Within a learning cycle $t$, KAqN learns classification tasks one at a time, using a dedicated single-label dataset for each task. %In learning iteration $i$, KAqN's learnable parameters are initialized with the parameter values learned from the preceding iteration, $\pmb{\theta}(i-1)$. 
In completing a learning iteration described in Eqs. (\ref{eq:Swin-B}, \ref{eq:projector}, \ref{eq:classifier}, \ref{eq:total_loss}) , KAqN's parameter values are updated:
\begin{equation}
    \pmb{\theta}(i) \xleftarrow{\text{monotask learning}}\pmb{\theta}(i-1),
    \label{eq:update theta}
\end{equation}
and so the classifier for that task:
\begin{equation}
    \pmb{\phi}_m(t)\xleftarrow{\text{monotask learning}} \pmb{\phi}_m(t-1).
\end{equation}
$\pmb{\theta}$ in Eq. (\ref{eq:update theta}) is the collection of the encoder's parameters $\pmb{\theta}^{\mathcal{E}}$ and the projectors' parameters $\pmb{\theta}^{\mathcal{P}}$. 
At the end of learning cycle $t$, KAqN completes its training on all the $M$ tasks. The resulting parameter values , $\pmb{\theta}(Mt)$, embed the acquired new knowledge about the driving scene classification.

%\textbf{Knowledge Acquisition from Mono-task Classifications via Cyclic Training.} Leveraging a single model to learn multiple tasks from single-label datasets of the similar domain can solve model redundancy when using single-task method, and costly manual labeling when using multi-task methods.  \cite{ark} proposed a novel framework to address the heterogeneous annotation across various datasets by cyclically training a teacher-student model. Like the function of the hippocampus in the human and animal brain, cyclically learning from diverse tasks plays a critical role in the acquisition of knowledge\cite{eichenbaum2017prefrontal}. This approach effectively circumvents accidental gradient explosions and convergence challenges that often arise in concurrent training scenarios. Cyclically training each individual mono-classification task fuses and integrates feature information and within a shared backbone, with each task allocated a corresponding classifier. All classifiers share the same structure, they differ in the number of output classes. The predicted outputs $\hat{y}_{i,j}$ for classification task is denoted as:

\iffalse
  \begin{equation}
 \hat{y}_{i,j} = \operatorname{FC}(\operatorname{MLP}(\operatorname{GAP}(fm_\textsc{iv});z_{i,j}^{n-1}); z_{i,j}^n),
  \end{equation}
\fi

\subsubsection{Knowledge Accumulation}

Knowledge acquired by KAqN in the cyclical training process accumulates in KAcN. Fig. \ref{fig:methodology} shows that KAcN's network architecture is identical to KAqN. The feature map extracted in each stage of the KAcN's encoder is $\pmb{f}^{'}_s$, which is further embedded as a feature vector $\pmb{d}^{'}_s$, for $s=\textsc{i},\dots,\textsc{iv}$. For simplicity, $\pmb{\theta}^{'}$ designates all the learnable parameters of KAcN.
The identical network architecture of KAcN and KAqN  greatly simplifies the process of knowledge accumulation delineated as below. 

Upon the completion of the learning cycle $t$, KAA needs to consolidate the new knowledge acquired by KAqN, $\pmb{\theta}(Mt)$, with the previously learned knowledge stored in KAcN, $\pmb{\theta}^{'}(t-1)$. For this consolidation, KAA adopts exponential moving average (EMA), a temporal mechanism for accumulating knowledge from cyclical learning  \cite{tarvainen2017mean}: 
\begin{equation}
\pmb{\theta}'(t)=\alpha(t)\pmb{\theta}'(t-1)+(1-\alpha(t))\pmb{\theta}(Mt),
\label{eq:accumulation}
\end{equation}
where $\alpha(t)$ is the stability coefficient, a real-valued parameter within the range $[0.9, 1]$. $\alpha(t)$ determines the proportion of previously learned knowledge to retain in cycle $t$. 

KAA, as a learning system, needs to keep a balance between retaining already learned knowledge and acquiring new knowledge, a phenomenon called stability-plasticity tradeoff in neuroscience \cite{neuron}, incremental learning \cite{incremental}, and continual learning \cite{continual}. If KAA is too stable, it will struggle to learn new knowledge effectively. Conversely, if it is too plastic, it risks forgetting important knowledge that has already been acquired. This paper designs a cosine scheduler for progressively updating the stability coefficient $\alpha(t)$ in Eq. (\ref{eq:accumulation}) from 0.9 to 1 in the cyclical learning process:
\begin{equation}
\alpha(t) = 0.9+0.05 \left[ 1 - \cos \left( t\pi/T \right) \right],
\label{eq:alpha_t}
\end{equation}
where $T$ is the maximum learning cycles for KAA. Eq. (\ref{eq:alpha_t}) indicates that $\alpha(t)$ is a monotonically increasing function. That is, in the early phases of learning, KAA leans toward plasticity, allowing itself to rapidly absorb new knowledge. In the later phases, it prioritizes stability to prevent the disruption of the established cognitive capability.

%consolidating existing knowledge to prevent excessive interference from new information that could disrupt the established cognitive framework. 
%To integrate and accumulate knowledge obtained from various single-task classifications, we adopt an exponential moving average (EMA) \cite{tarvainen2017mean} to incrementally update the weights of teacher model. Concretely, the EMA update is a temporary sequence mechanism, which can accumulate historical knowledge derived from cyclical training. We denote the weight update for the Knowledge Accruing Network as:

%where $\theta_t^{(\text{teacher})}$ and $\theta_t^{(\text{student})}$ is the weight of teacher-student model respectively, $t$ is the epoch-wised training step, and $\beta$ is coefficient schedule parameter. This concept is similar to the "stability-plasticity tradeoff" in neuroscience\cite{neuron}, incremental learning\cite{incremental}, and continual learning\cite{continual}. In the early stages of learning, our brain leans toward plasticity, allowing us to rapidly absorb and adjust to new knowledge. In the later stages, however, the brain prioritizes stability, consolidating existing knowledge to prevent excessive interference from new information that could disrupt the established cognitive framework. Therefore, $\beta_t$ is determined using a cosine schedule from 0 to 1, and it is defined as:

\subsubsection{Knowledge Retention} 
\label{subsubsec:Knowledge Reinforcement}
%\textbf{Reinforce Memory via Multilayer Consistency Loss.} 

The cyclical nature of training can cause catastrophic forgetting, thus impeding continual learning and improvement. This issue can be effectively addressed by setting the accumulated knowledge as the reference for the subsequent knowledge acquisition. KAA achieves this through feature-based KD, which penalizes large stage-wise discrepancies between the feature embedding extracted by KAqN, $\pmb{d}_s(i)$, in each learning iteration and the corresponding feature embedding derived by KAcN, $\pmb{d}'_s(t-1)$. The penalty is calculated as the mean squared error (MSE):
\begin{equation}
    \mathcal{L}_{cst,s} (i)= \frac{1}{|\pmb{d}_s|}\|\pmb{d}_s (i)-\pmb{d}^{'}_s (t-1)\|^2_2,
\end{equation}
for stages $s = \textsc{I}, \dots, \textsc{IV}$. This particular KD design fully leveraging the advantages of deep networks, which has been shown to be effective  \cite{romero2014fitnets}.

Then, the overall consistency loss is the weighted sum of stage-wise consistency losses:
\begin{equation}
 \mathcal{L}_{cst} (i)  = \sum_{s=\textsc{i}}^{\textsc{iv}} \lambda_s\mathcal{L}_{cst,s}(i) ,
 \label{eq:loss_consistency}
\end{equation}
where $\lambda_s$ is the coefficient for stage $s$, which is set to be one for all stages in this study. 

\subsubsection{Loss Function as the Learning Guidance} 
\label{subsubsec:Loss Function}

The consistency loss in Eq. (\ref{eq:loss_consistency}) is a regularization term for training KAqN, ensuring that the new knowledge acquisition is based on already attained knowledge. Therefore, the per-sample loss function in learning iteration $i$ can be formulated as the weighted sum of the prediction loss and the consistency loss:
\begin{equation}
    \mathcal{L}_{tot} (i)= \beta (i) \mathcal{L}_{cls} (i) + (1-\beta(i))\mathcal{L}_{cst} (i),
    \label{eq:total_loss}
\end{equation}
where $\beta(i)$, a real-valued parameter in the range $[0,1]$, is the performance-based acquisition-retention indicator (PARI) for balancing the acquisition of new knowledge and the retention of already attained knowledge. $\beta(i)$ is adjusted dynamically with respect to learning needs. Lower performance anticipated for KAqN on the task in learning corresponds to a greater need for continual learning and improvement on this task. Therefore, $\beta(i)$ is designed as a monotonically decreasing function of the estimated task performance, $\widehat{p}(i)$:
\begin{equation}
    \beta(i) = (1-\widehat{p}(i)^\psi)^{1/\psi},
    \label{eq:beta(i)}
\end{equation}
where $\psi>1$ is a real-valued parameter defining the shape of the curve, which is set to be 4 in this study. 

$\widehat{p}(i)$ in Eq. (\ref{eq:beta(i)}) estimates classification accuracy of KAqN in learning iteration $i$, obtained based on an EMA process:
\begin{equation}
    \widehat{p}(i) =\omega\widehat{p}(i-M) + (1-\omega) p(i-M),
    \label{eq:p(i)}
\end{equation}
where $\widehat{p}(i-M)$ is the estimated accuracy of KAqN for the last learning cycle and $p(i-M)$ is the accuracy that KAqN actually achieved. The EMA process ensures that the performance estimation is based on the entire record from past training cycles. %When a prior measurement or estimation is not available, a default value, $\widehat{p}(0)$, representing KAqN's initial (low) performance, is used as a substitute. 
The parameter $\omega$ is set to be $0.9$ in this study, meaning that the performance estimation puts more weight on the long-term trend than the most recent observation.

Summing the loss function in Eq. (\ref{eq:total_loss}) over all the training data yields the total training loss. By minimizing this loss with respect to KAqN's learnable parameters in each learning iteration and across learning cycles, the classification model defined in Eq. (\ref{eq:classification model}) is optimized.
The resulting classification model $\widetilde{C}$,  described in Algorithm \ref{alg:KAAN}, is a foundation model for driving scene identification.

\subsection{Consistency-based Active Learning (CAL)}\label{subsec:CAL}

KAA learns the knowledge for multi-label scene classification through monotask learning across different single-label datasets. While simplifying data collection and reducing learning complexity, this approach confronts a domain shift issue. That is, the class label distributions of single-label datasets represent the marginal probability distributions of individual attributes, different from their joint distribution. The same issue occurs in the extracted features. Therefore, the foundation model can make mistakes when it attempts to simultaneously identify a scene using multiple attributes.

\begin{figure}[b]
    \centering
    \includegraphics[width=\linewidth]{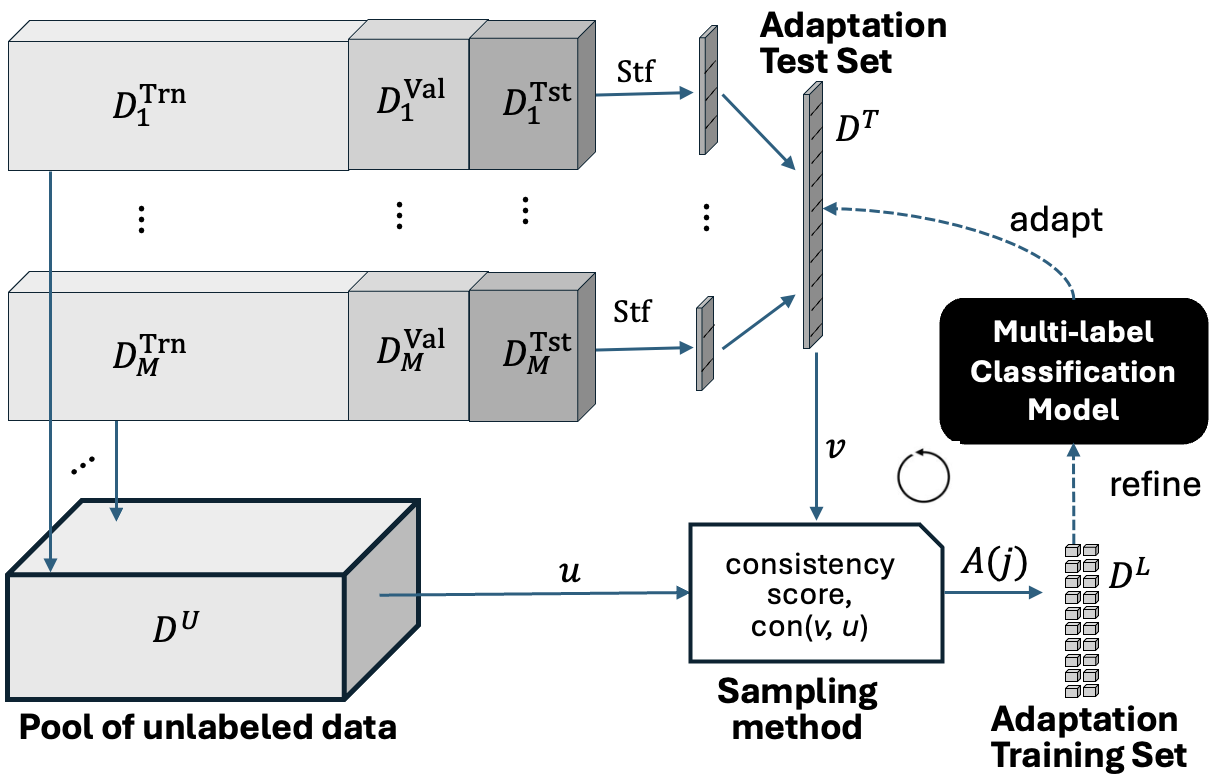}
    \caption{Schematic diagram for Consistency-based Active Learning (CAL)}
    \label{fig:CAL-data}
\end{figure}

\begin{algorithm}
\SetKwInput{In}{INPUT}
\SetKwInput{Init}{INITIALIZATION}
\SetKwInput{CAL}{CONSISTENCY ACTIVE LEARNING}
\caption{CAL Algorithm}
\label{alg:alg_CAL}
\In{\\
  $\mathcal{D}^U$: Unlabeled pool\\%
  $\mathcal{D}^T$: Test dataset\\%
  $b_j$: Budget for adaptation iteration $j$\\%
  $N_{\text{CAL}}$: Maximum iterations%
}
\BlankLine
\Init{\\
  $\mathcal{D}^L = \emptyset$: initial labeled training set\\
  $\pmb{\Theta}^{'}(0)=\widetilde{\pmb{\Theta}}$: initial learnable parameters 
}
\BlankLine
\CAL{\\
\For{$j = 1, \dots, N_{\text{CAL}}$}
{
  compute consistency score $\mathrm{con}(v,\overline{u})$, $\forall\,v \in \mathcal{D}^T$;\\
  
  $\mathcal{A}(j) $ comprises $b_j$ samples from $\mathcal{D}^U$ with top consistency scores, labeled by the Oracle;\\
  
  add $\mathcal{A}(j)$ to the labeled dataset: \\
  \Indp
  $\mathcal{D}^L \leftarrow \mathcal{D}^L \,\cup\, \mathcal{A}(j)$;\\
  \Indm
  
  remove $\mathcal{A}(j)$ from unlabeled dataset:\\
  \Indp
  $\mathcal{D}^U \leftarrow \mathcal{D}^U \setminus \mathcal{A}(j)$;\\
  \Indm
  
  train the classification model for multitasking: \\
  \Indp
  $\pmb{\Theta}^{'}(j)\xleftarrow{\ \text{multitask learning}\ } \pmb{\Theta}^{'}(j-1)$.\\
  \Indm
  \BlankLine
  \If{stopping criterion is satisfied}{%
    $j^* = j$.\\
    \textbf{break}
  }
}
}
\BlankLine
\KwOut{%
  $\pmb{\Theta}^\ast=\pmb{\Theta}^{'}(j^\ast)$
}
\end{algorithm}

This paper proposes CAL (Algorithm \ref{alg:alg_CAL}) to tackle the aforementioned issue, which seeks to cost-effectively adapt the attained foundation model to real-world multi-attribute scene identification. Fig. \ref{fig:CAL-data} illustrates the overall approach of the proposed CAL.
 A multi-label test sample, $\mathcal{D}^T$, representing the driving scene distribution in the joint space of scene attributes, is prepared using a stratified sampling strategy ($\operatorname{Stf}$). It randomly samples $\kappa$ test images per class from DSI's test sets: 
\begin{equation}
    \mathcal{D}^T = \cup_{m=1}^M \operatorname{Stf}(\mathcal{D}_m^{\text{Tst}};\mathcal{U}_m, \kappa),
\end{equation}
which is then fully annotated with respect to all $M$ attributes. The adaptation test set may also be drawn from alternative domains or tasks to which the foundation model has been adapted. The adaptation test set may also be drawn from other domains or tasks to which the foundation model has been adapted.

Then, a training set $\mathcal{D}^L$ needs to be labeled for adapting the classification model to multitask learning over iterations. Although the ground truth labels for the training set $\mathcal{D}_m^{\text{Trn}}$ were seen by task $m$, other tasks have never seen these labels. Therefore, the union of DSI's training sets forms an ``unlabeled'' pool,
\begin{equation}
    \mathcal{D}^U=\cup_{m=1}^M \mathcal{D}_m^{\text{Trn}},
    \label{eq:D^T}
\end{equation}
from which training samples can be selected and prepared according to Algorithm \ref{alg:alg_CAL}. 

CAL searches the unlabeled pool $D^U$ for data points that are similar to those in the test set $D^T$. A consistency function, $\operatorname{con} (v, u)$, is defined to quantify the similarity between any image in the test set, $v\in\mathcal{D}^T$, and any one in the unlabeled pool, $u\in\mathcal{D}^U$:
\begin{equation}
    \operatorname{con}(v,u) = -\|v,u\|_{\pmb{d}^{'}_{2,\textsc{iv}}},
    \label{eq:consistency}
\end{equation}
where $\|\cdot,\cdot\|_{\pmb{d}^{'}_{2,\textsc{iv}}}$ measures the Euclidean distance between two data points in terms of their final-stage feature embeddings. Thereby, for any image in the test set $v\in\mathcal{D}^T$, the corresponding image in the unlabeled pool which is most similar to it is identified:
\begin{equation}
    \overline{u} = \argmax_{u\in\mathcal{D}^U}\operatorname{con} (v, u).
\end{equation}
Then, these unlabeled images are ranked in decreasing order of their consistency score:
\begin{equation}
    \operatorname{con}(v^{(1)},\overline{u}^{(1)})>\operatorname(v^{(2)},\overline{u}^{(2)})>\dots
\end{equation}
to prioritize them for annotation. Following the ranking, the top $b_j$ unlabeled images with the highest consistency scores are selected for annotation:
\begin{equation}
    \mathcal{A}(j) =\{\overline{u}^{(1)},\dots, \overline{u}^{(b_j)}\},
\end{equation}
where $b_j$  is the annotation budget in iteration $j$. 
After that, $\mathcal{A}(j)$ is moved from the unlabeled pool to the labeled training set, $\mathcal{D}^L$:
\begin{equation}
    \mathcal{D}^L \leftarrow \mathcal{D}^L\cup \mathcal{A}(j),
\end{equation}
\begin{equation}
    \mathcal{D}^U \leftarrow \mathcal{D}^U \setminus \mathcal{A}(j).
\end{equation}
The dataset $\mathcal{D}^L$ is used to refine the classification model into a multitask model.  

In refining the classification model, the focal loss is evaluated for each data point in the training set $\mathcal{D}^L$ to focus on hard samples for each task, 
\begin{equation}
    \mathcal{L}_{MT} = - \sum_{m=1}^M \pmb{y}_m\cdot((\pmb{1}-\widehat{\pmb{y}}_m)^{\gamma_m}
\odot \log\widehat{\pmb{y}}_m),
\label{eq:focal loss}
\end{equation}
where $\odot$ stands for the element-wise product,  $\cdot$ is the dot product, and $\gamma_m$ is the focusing parameter for task $m$, which is set to 1 here. The total loss of the multitask learning process is obtained by summing up this loss across in Eq. (\ref{eq:focal loss}) across all data points in $\mathcal{D}^L$, which guides the model refinement iteratively:
\begin{equation}
\pmb{\Theta}^{'}(j)\xleftarrow{\text{multitask learning}} \pmb{\Theta}^{'}(j-1),
\end{equation}
where $\pmb{\Theta}^{'}(0) =\widetilde{ \pmb{\Theta}}$. This process can go for several iterations until satisfaction, resulting in the optimized classification model $\mathcal{C}^{\ast}$ parameterized with $\pmb{\Theta}^{\ast}$.

\section{Implementation Details}
\label{sec:Implementation Details}

The proposed KAA-CAL learning method was implemented using PyTorch 1.10.0 on a server equipped with an Nvidia Tesla V100 featuring 32 GB of memory. Model training utilized the AdamW optimizer, a batch size of 32, and a maximum of 100 training epochs. Input images were resized to $224\times244$ and augmented with random rotations, crops, and pixel normalization.

%\subsection{KAA Implementation} 
The deep encoder for KAA in Eq. (\ref{eq:Swin-B}) was initialized with the weights pretrained on ImageNet-22k \cite{imagenet}. The projectors in Eq. (\ref{eq:projector}) and the classifiers in Eq. (\ref{eq:classifier}) were initialized using the default Xavier initialization method. Then, learnable parameters were progressively optimized using the method delineated in Algorithm \ref{alg:KAAN}. The learning rate for KAA implementation was adjusted in two phases. The first phase is a 10-epoch warm-up, where the learning rate linearly increased from an initial value of 1e-6 to 5e-4. In the second phase, the learning rate decayed to a final value of 1e-5 using a cosine schedule. Upon convergence at the end of all training epochs, the model from the final epoch is the foundation model.
%\cite{Xavier}

%The hyperparameters for the learning rate (lr) schedule include a 10-epoch warm-up phase, starting with a default lr of 1e-6, increasing to a starting lr of 5e-4, and ending at 1e-5 with a cosine schedule. The weights are loaded from a pretrained model on ImageNet-22K \cite{imagenet} and are strictly consistent with those provided in the original paper. %The coefficient $\lambda_i$ of four-stages consistency loss are set equally to 1, the $n$ in Function \ref{AAPI} is set to 4.

%\subsection{CAL Implementation} 

In implementing CAL, a per-iteration budget ($b_j$) of 200 images, approximately 1\% of the single-label training data in DSI, was allocated for annotating the data recommended by Algorithm \ref{alg:alg_CAL}. This process ran for up to five iterations to evaluate the efficiency of CAL. The data annotation process confronted a challenge - not all images can be reliably annotated with all the seven driving scene attributes. For example, annotators found the weather condition in some nighttime images was hard to identify. When a reliable label was unavailable, a sentinel value ``-1'' was assigned. Such labels were excluded from the loss function evaluation to ensure training integrity. Aiming to retain knowledge learned from KAA, the first three stages of the deep encoder were frozen for finetuning multitask classification. The learning rate scheduler for the CAL implementation was similarly configured.%it began with a 10-epoch warm-up, with the learning rate linearly increasing from 1e-6 to 1e-3. Then, the learning rate decayed to a final value of 1e-5 based on a cosine schedule.  

\section{Experiments and Results}\label{resultandexper}

Experimental studies were conducted, aiming to verify  the merits of the proposed KAA-CAL learning method.

\subsection{Deep Encoder Selection}
\label{subsec:deep encoder selection}
Various deep network architectures are available to choose from for the deep encoder. Vision transformers (ViT) demonstrate unique advantages over convolutional neural networks  in learning deep features for image classification. This study compared SOTA transformer-based models, including Swin-S, Swin-B, ViT-B, and ViT-L. The comparison spanned the seven independent image classification tasks, evaluating models from the perspectives of both classification accuracy and model complexity. The results are summarized in TABLE \ref{tab:performance}.%, with the highest accuracy achieved on each task being bolded.

\begin{table}[!ht]
    \caption{Classification accuracy (\%) comparison of vision transformers as the deep encoder in monotask learning }\label{tab:performance}
    \centering
        \begin{tabular}{l|>
        {\raggedleft\arraybackslash}p{0.34in}>{\raggedleft\arraybackslash}p{0.34in}>{\raggedleft\arraybackslash}p{0.28in}>
        {\raggedleft\arraybackslash}p{0.28in}}
       
            \toprule
            & Swin-B & Swin-S & ViT-B & ViT-L \\\midrule
            {\textbf{MODEL COMPLEXITY}}&&& \\
            FLOPs (G) & 15.4 & 8.7 & 17.6 & 76.9 \\
            The number of parameters (M) & 88 & 50 & 86 & 307 \\\midrule
            {\textbf{ACCURACY} (\%)}&&& \\
            Time of Day & \textbf{99.6} & 91.3 & 89.3 & 96.0 \\
            Weather & \textbf{92.2} & 92.0 & 90.9 & 92.1 \\
            Weather-related Road Condition & \textbf{98.4} & 97.7 & 96.7 & 96.6 \\
            Road Function & \textbf{99.8} & 99.2 & 95.4 & 95.8 \\
            Intersection & 88.6 & 86.9 & 93.4 & \textbf{95.8} \\
            Work Zone & \textbf{95.2} & 87.9 & 88.5 & 90.2 \\
            Grade-separated Infrastructure & 94.6 & 93.7 & 97.2 & \textbf{98.4} \\
            %\midrule
            \hline
            
            Average & \textbf{95.5} & 92.7 & 93.4 & 95.0 \\\bottomrule
        \end{tabular}
        \raggedright
        The top performance for each task is bolded.
\end{table}

As expected, Swin-B outperforms Swin-S on all seven tasks, and the same pattern holds for ViT-L compared to ViT-B. This confirms that the more complex transformers (Swin-B and ViT-L) achieve better performance than their less complex counterparts. While both are powerful, Swin transformers address limitations of the original ViT, exhibiting improved computational efficiency and the capability for multi-scale feature learning. TABLE \ref{tab:performance} shows that Swin-B outperformed ViT-L on five out of seven tasks with a margin ranging from 0.1\% to 4.9\%. Importantly, the FLOPs for Swin-B are only 20\% of ViT-L. Given the highest average accuracy and more affordable model complexity, Swin-B is chosen as the deep encoder architecture for the classification model.   

\subsection{Effectiveness of Stage-wise Knowledge Distillation}
\label{subsection:Effectiveness of Stage-wise Knowledge Distillation}

The ability to retain previously acquired knowledge while learning new information is essential for the multi-attribute scene identification model to be scalable. Sections \ref{subsubsec:Knowledge Reinforcement} and \ref{subsubsec:Loss Function} explained that, when KAqN continues acquiring new knowledge, the already acquired knowledge accumulated in KAcN is retained through stage-wise feature-based KD, and the intensity of this regularization is moderated by PARI. To demonstrate the effectiveness of this knowledge retention design for KAA, an ablation study about KD was performed with results summarized in TABLE  \ref{tab:compo}.

\begin{table}[b]
    \caption{Effectiveness of Knowledge Retention through Stage-wise Feature-based Knowledge Distillation}\label{tab:compo}
    \centering
        \begin{center}
        \begin{tabular}{l|>
        {\centering\arraybackslash}p{0.5in}>{\centering\arraybackslash}p{0.5in}>{\centering\arraybackslash}p{0.3in}>
        {\centering\arraybackslash}p{0.5in}}
            %\hline
            \toprule
             Model &  Last-stage KD & Stage-wise KD & PARI & Avg. (\%) \\\midrule
            Baseline & &    &  & 92.7 \\
            Model \textsc{I}&\checkmark &    &  & 92.8 \\
            Model \textsc{II}&\checkmark &    & \checkmark &  94.0\\
            Model \textsc{III}& & \checkmark &  & 94.5 \\
            Model \textsc{IV} (Ours)& & \checkmark & \checkmark  & \textbf{95.2} \\\bottomrule
        \end{tabular}
        \end{center}
\end{table}

The Baseline model in TABLE \ref{tab:compo} is learned solely by a student network because both stage-wise KD and PARI are removed. It results in an average accuracy  of 92.7\%, representing a 2.5\% decrease compared to our model (Model IV). This gap will increase when the KAA attempts to acquire more diverse knowledge from many datasets.

Model \textsc{I} adds the last-stage KD to the baseline model, analogous to a teacher who provides less guidance to the student's learning. This model achieves a negligible performance gain of merely 0.1\%. Conversely, Model \textsc{III} incorporates stage-wise KD into the baseline model, leading to an average accuracy increase of 1.8\%. This comparison highlights the importance of stage-wise KD, since features of different scales offer comprehensive information about driving scenes.    

Model \textsc{II} augments the last-stage KD with PARI, which improves average accuracy by 1.2\% over Model \textsc{I}. Similarly, Model \textsc{IV} pairs PARI with stage-wise KD, increasing Model \textsc{III}'s well-performed average accuracy by an additional 0.7\%. These observations verify the importance of dynamically adjusting the weight for new knowledge acquisition according to specific learning needs, regardless of the KD method.

\subsection{Performance of the Foundation Model}
\label{subsection:Performance of the Foundation Model}

\begin{figure}[!ht]
  \centering
  \includegraphics[width=\columnwidth]{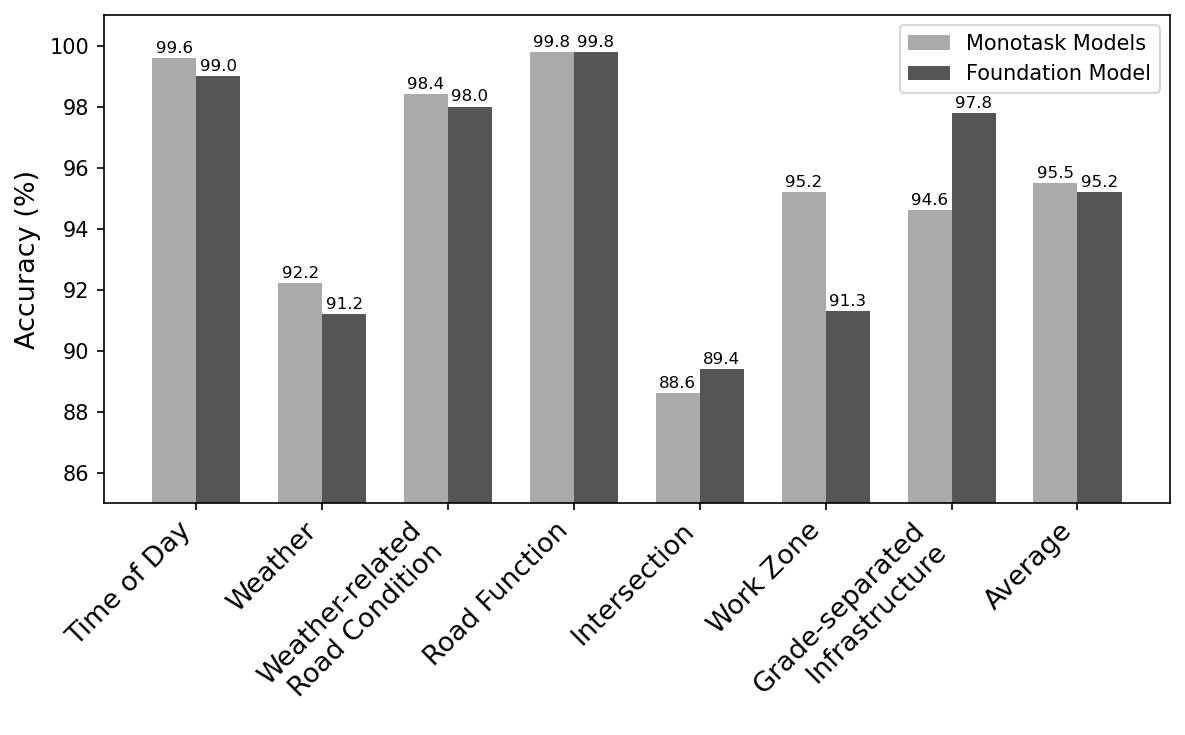}
  \caption{The foundation model vs. corresponding monotask models}\label{fig:result1}
\end{figure}

After verifying the strengths of KAA's learning architecture design, the experimental study continued to assess the resulting foundation model's performance. Fig. \ref{fig:result1} compares the foundation model to the seven individual monotask models. As illustrated, the foundation model is comparable to the  individual monotask models in task accuracy, with variations ranging from -3.9\% to 3.2\% across the seven tasks. Specifically, the foundation model is 3.2\% more accurate than the monotask model for identifying Grade-separated Infrastructure. However, the foundation model's accuracy in identifying Work Zone is 3.9\% lower. This drop can be attributed to the challenge of image-level classification in capturing the fine-grained features critical for identifying work zones, such as safety cones and work zone signs. Approaches based on object detection or semantic segmentation (e.g., \cite{work1}) are generally more capable of detecting such localized features. For the remaining five tasks, their performance differences are within $\pm 1$\%. The foundation model can substitute for individual models dedicated to different classification tasks, as its comparable performance demonstrates.
%work2

\subsection{Effectiveness of Consistency-based Active Learning}
\label{subsection:Effectiveness of Consistency-based Active Learning}
%\subsubsection{Comparison with SOTA Deep Active Learning Methods} 

\begin{figure*}[b]
  \centering
  \includegraphics[width=6.8in ]{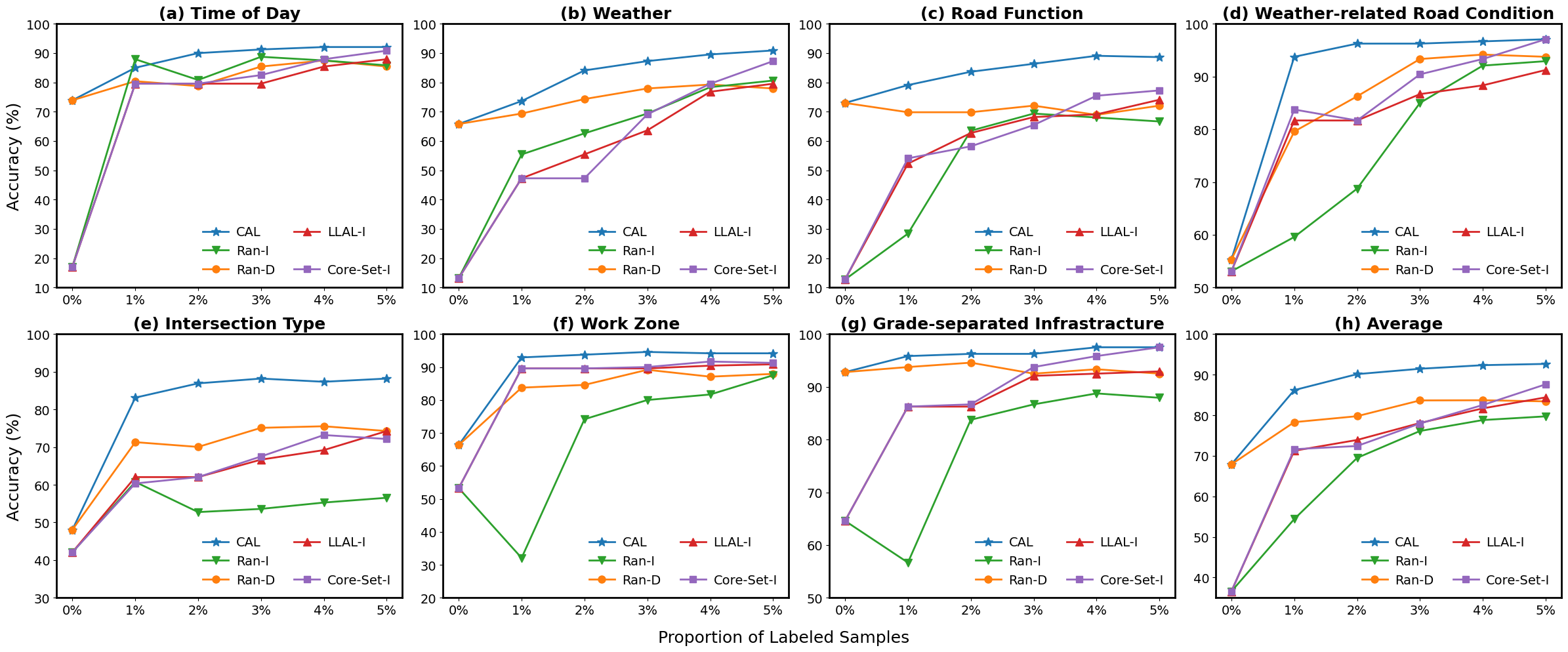}
  \caption{Comparison of active learning methods in terms of task and average accuracies (\%)}\label{fig:result2}
\end{figure*}

The foundation model, obtained from diverse single-label datasets via monotask learning, must quickly adapt to downstream tasks. CAL proposed in Section \ref{subsec:CAL} is designed for this purpose, addressing the limitations inherent in KAA's monotask learning approach. To demonstrate its effectiveness, CAL was compared to its variants and SOTA deep active learning methods, detailed as follows.

\begin{itemize}
    \item \textbf{Ran-I}'s deep encoder is initialized with the weights pretrained on ImageNet-22k. Samples for each class are randomly selected from the unlabeled pool and then annotated. Compared to CAL, Ran-I does not leverage the driving scene knowledge offered by the foundation model, nor does it employ the feature consistency-based criterion for sampling.
    \item \textbf{Ran-D} uses the same sampling method as Ran-I, but it adopts the foundation model as its deep encoder. 
    \item \textbf{Core-Set-I} is a diversity-based active learning approach proposed in \cite{coreset}. Its deep encoder is initialized with the weights pretrained on ImageNet-22k. In the first iteration, the model is refined with an initial training set, helping expose the encoder to some multi-label training data. In each subsequent iteration, a k-Center-Greedy strategy selects additional samples to represent various clusters in the unlabeled pool.
    \item \textbf{LLAL-I} follows the same initialization as Core-Set-I, but it introduces an additional loss prediction component \cite{learningloss}, which learns to  estimate a pseudo-loss value for each sample. In selecting samples from the unlabeled pool, those with the highest pseudo-loss values are prioritized, as they represents the most informative samples to learn from.
\end{itemize}

Fig. \ref{fig:result2} compares the proposed CAL method with those introduced above, presenting the task-level accuracies in plots a-g and the average accuracy in plot h. Each plot illustrates the test accuracy values before adaptation (the training set $\mathcal{D}^L$ has 0\% labeled data from $\mathcal{D}^U$) and over five iterations of adaptation. Each iteration selects 1\% of data from the unlabeled pool. Fig. \ref{fig:result2}(h) shows that, prior to adaptation, models that are initialized with the foundation model's weights (i.e., CAL and Ran-D) demonstrate a higher average accuracy than those initialized with weights pretrained on ImageNet (i.e., Ran-I, LLAL-I, and Core-set-I), with a margin of 31.3\%. The initial performance gap between these two groups highlights the benefit of utilizing the foundation model for driving scene identification, as it possesses the knowledge about the seven classification tasks.

After the first iteration, CAL is still the top one although all methods improve their task performance. It surpasses the best competing method's accuracy by a significant margin of 3.3\% to 10.0\% across all tasks, except for the task of identifying Time of Day (Fig. \ref{fig:result2}(a)). Effectively, CAL brings the classification model's average accuracy up to 86.2\% after just one iteration of adaptation. With another iteration, CAL boosts the average accuracy to 90.1\%, with task accuracy ranging from 83.6\% to 96.3\%. Through additional iterations, CAL further enhances the accuracy for identifying Weather and Road Function, thus increasing the average accuracy to 92.7\%. Figure \ref{fig:result2}(h) reveals CAL's superior performance and cost-effectiveness. It consistently outperforms the best competing method in every iteration, with margins ranging from 5.0\% to 12.9\%. Importantly, CAL has achieved its major improvements in just two iterations, while all other methods require significantly more iterations and so more training data for adaptation. 

The importance of CAL's sample recommendation method is demonstrated by a comparison between Ran-D and CAL, which differ in their sampling strategies. Ran-D uses a simple stratified random sampling strategy, whereas CAL samples data based on features consistency measurement between unlabeled data and the test set for adaptation. CAL consistently improves task accuracy over iterations, whereas Ran-D lowers the accuracy in identifying Road Function and Grade-separated Infrastructure. Moreover, CAL achieves a higher saturated accuracy than Ran-D on all tasks. Consequently, CAL dominates Ran-D in Fig. \ref{fig:result2}(h).

The groundwork that the driving scene foundation model has laid for multi-attribute scene identification is highlighted by reviewing Ran-I, LLAL-I, and Core-Set-I that do not utilize the foundation model. Those methods need significantly more iterations to improve task performance. Among those, Core-Set-I is the most competitive. After five iterations, it has attained the task accuracy comparable to CAL's in identifying Weather-related Road Condition and Grade-separated Infrastructure. But CAL still beat Core-Set-I on the remaining five tasks by a margin ranging from 1.3\% to 13.9\%. Consequently,  Core-Set-I's average accuracy is still 11.8\% lower than that of CAL after three iterations and 5.0\% lower after five iterations. 

\begin{table*}[t]
    \caption{Progressive improvement in task and average accuracies (\%) in the ablation study}\label{tab:abla1}
    \centering
        \begin{tabular}{l|cc|lllllll|l}
            \toprule
              Model &KAA & CAL&ToD&Wea&RoF&WrR&Int&WrZ&GsI& Average\\
              \midrule
            Baseline & & &17.1&13.2&12.7&53.1&42.1&53.3&64.6&36.6 \\
            Foundation &\checkmark & &73.9 \tiny{56.8$\uparrow$}&65.8 \tiny{52.6$\uparrow$}&73.0 \tiny{60.3$\uparrow$}&55.3 \tiny{$\;$2.2$\uparrow$}&48.1 \tiny{$\;$6.0$\uparrow$}&66.4 \tiny{13.1$\uparrow$}&92.8 \tiny{28.2$\uparrow$}&67.9 \tiny{31.3$\uparrow$} \\
            with Adaptation&\checkmark & \checkmark (2 iters) &90.0 \tiny{16.1$\uparrow$}&84.1 \tiny{18.3$\uparrow$}&83.6 \tiny{10.6$\uparrow$}&96.3 \tiny{41.0$\uparrow$}&86.9 \tiny{38.8$\uparrow$}&93.8 \tiny{27.4$\uparrow$}&96.3 \tiny{ 3.5$\uparrow$}&90.1 \tiny{22.2$\uparrow$}\\
            with Adaptation&\checkmark & \checkmark (5 iters) &\textbf{92.1} \tiny{$\;$2.1$\uparrow$}&\textbf{90.9} \tiny{$\;$6.8$\uparrow$}&\textbf{88.6} \tiny{$\;$5.0$\uparrow$}&\textbf{97.1} \tiny{$\;$0.8$\uparrow$}&\textbf{88.2} \tiny{$\;$1.3$\uparrow$}&\textbf{94.2} \tiny{$\;$0.4$\uparrow$}&\textbf{97.5} \tiny{$\;$1.2$\uparrow$}&\textbf{92.7} \tiny{$\;$2.6$\uparrow$}\\
            \bottomrule
        \end{tabular}
        \vspace{0.5em}
        
        \parbox{0.88\linewidth}{\footnotesize ToD: Time of Day; Wea: Weather; RoF: Road Function; WrR: Weather-related Road Condition; Int: Intersection type; WrZ: Work Zone, GsI: Grade-separated Infrastructure.}
\end{table*}

The comparison in Fig. \ref{fig:result2} indicates 
that CAL benefits from the foundation model, which not only sets up a higher starting point for adaptation but also gives CAL good knowledge to identify the most useful training data for ongoing learning and refinement. This pairing of a capable active learning method with a knowledgeable foundation model is the underlying reason for the efficiency and effectiveness of model adaptation.

\subsection{Ablation Study} 
\label{subsection:ablation study}

KAA and CAL have now been verified as advantageous designs: KAA builds a knowledge foundation for driving scene identification by acquiring and accumulating information from heterogeneous single-label source data via monotask learning. CAL adapts this foundation model to the multi-label target data. An ablation study was conducted to gauge their respective contributions, with results summarized in TABLE \ref{tab:abla1}.

The baseline model in this ablation study is a multitask classification model whose deep encoder is Swin-B pretrained on ImageNet. This baseline model achieves task accuracies ranging from 17.1\% to 64.4\%, resulting in an average accuracy of 36.6\%. Its low accuracy indicates that ImageNet possesses insufficient knowledge regarding driving scene identification.

Adoption of the foundation model as the deep encoder boosts the average accuracy to 67.9\%, representing a substantial increase of 31.3\%. This gain is attributed to the comprehensive knowledge regarding driving scenes acquired by KAA from diverse single-label datasets. For instance, the accuracy in identifying Road Function, a unique attribute specific to driving scenes, had an increase of 60.3\%. Furthermore, the task accuracies for identifying the environmental attributes, Time of Day and Weather, each rise by over 50\%. Consequently, the task accuracies now span from 48.1\% to 92.8\%. However, room for improvement are clearly present. The gains in identifying Weather-related Road Condition and Intersection are limited, primarily due to a large shift in driving scene data distribution compared to the marginal distributions in the single-labeled datasets.

CAL, designed to address the limitation of KAA's monotask learning approach, effectively improves the average accuracy to 90.1\% with only two iterations, corresponding to another increase of 22.2\%. At the task level, accuracies have reached the range from 83.6\% to 96.3\%. With three additional iterations, the task accuracies in identifying Weather and Road Function are improved by at least 5\%, further boosting the average accuracy to 92.7\%.

\subsection{Driving Scene Identification Examples}
\label{subsection:qualitative examples}

\begin{table*}[t]

    \centering
    \caption{Groundtruth and predicted labels for driving scene identification examples}
    \label{tab:ablat2}
    \begin{tabular}{|c|c|l|l|l|l|l|}
        \hline
        \multirow{2}{*}{Example}&\multirow{2}{*}{Driving scene}&\multirow{2}{*}{Groundtruth}&\multicolumn{4}{c|}{Predicted Labels}\\
        \cline{4-7}
        &&&Baseline&Foundation&KAA-CAL (2 iters)&KAA-CAL (5 iters)\\\hline
        
        \multirow{7}{*}{a}&\multirow{7}{*}{\includegraphics[width=5cm, height=2.0cm, keepaspectratio]{nig8.png}}&Night&\textcolor{red}{Dawn}&Night&Night&Night\\
         &&x&x&x & x& x\\ 
         &&Interstate&\textcolor{red}{Collector}&\textcolor{red}{Local road} & Interstate& Interstate\\
         &&Dry&\textcolor{red}{Wet}&\textcolor{red}{Snowy} & Dry& Dry\\
         &&Non-intersection&Non-intersection&Non-intersection & Non-intersection& Non-intersection\\
         &&Non-work zone&\textcolor{red}{Work zone}&Non-work zone & Non-work zone & Non-work zone\\
         &&Open roads&Open roads&Open roads & Open roads& Open roads\\\hline
         
         \multirow{7}{*}{b}&\multirow{7}{*}{\includegraphics[width=5cm, height=2cm, keepaspectratio]{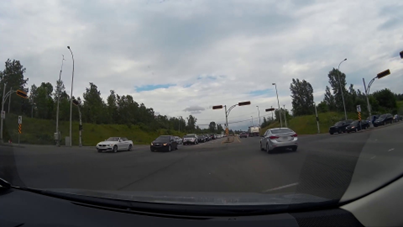}}&Daytime&\textcolor{red}{Dawn}&Daytime&Daytime&Daytime\\
         &&Overcast&\textcolor{red}{Snowing}&Overcast & Overcast& Overcast\\ 
         &&Arterial&\textcolor{red}{Collector}&\textcolor{red}{Local road}&\textcolor{red}{Local road} & Arterial\\
         &&Dry&\textcolor{red}{Wet}&\textcolor{red}{Snowy} & Dry& Dry\\
         &&4-way&\textcolor{red}{3-way}&4-way & 4-way& 4-way\\
         &&Non-work zone&\textcolor{red}{Work zone}&\textcolor{red}{Work zone} & Non-work zone& Non-work zone\\
         &&Open roads&Open roads&Open roads & Open roads& Open roads\\\hline

         \multirow{7}{*}{c}&\multirow{7}{*}{\includegraphics[width=5cm, height=2cm, keepaspectratio]{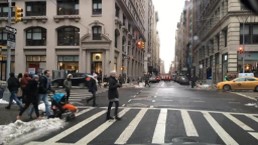}}&Daytime&\textcolor{red}{Dawn}&\textcolor{red}{Dawn}&Daytime&Daytime\\
         &&Overcast&\textcolor{red}{Snowing}&\textcolor{red}{Snowing} & Overcast& Overcast\\ 
         &&Local road&\textcolor{red}{Arterial}&Local road & Local road& Local road\\
         &&Snowy&Snowy&Snowy & Snowy& Snowy\\
         &&4-way&\textcolor{red}{Non-intersection}&4-way & 4-way& 4-way\\
         &&Non-work zone&\textcolor{red}{Work zone}&Non-work zone & Non-work zone& Non-work zone\\
         &&Open roads&Open roads&Open roads & Open roads& Open roads\\\hline

         \multirow{7}{*}{d}&\multirow{7}{*}{\includegraphics[width=5cm, height=2cm, keepaspectratio]{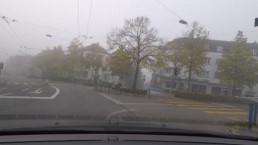}}&Daytime&\textcolor{red}{Dawn}&\textcolor{red}{Dawn}&Daytime&Daytime\\
         &&Foggy&\textcolor{red}{Snowing}&Foggy & Foggy& Foggy\\ 
         &&Local road&\textcolor{red}{Collector}&Local road & Local road& Local road\\
         &&Dry&\textcolor{red}{Snowy}&\textcolor{red}{Snowy} & Dry& Dry\\
         &&3-way&\textcolor{red}{Non-intersection}&\textcolor{red}{4-way} & 3-way& 3-way\\
         &&Non-work zone&\textcolor{red}{Work zone}&Non-work zone & Non-work zone& Non-work zone\\
         &&Open roads&Open roads&Open roads & Open roads& Open roads\\\hline

         \multirow{7}{*}{e}&\multirow{7}{*}{\includegraphics[width=5cm, height=2cm, keepaspectratio]{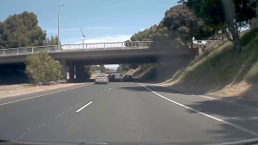}}&Daytime&\textcolor{red}{Dawn}&Daytime&Daytime&Daytime\\
         &&Clear&\textcolor{red}{Snowing}&\textcolor{red}{Overcast}&\textcolor{red}{Overcast} & Clear\\ 
         &&Interstate&\textcolor{red}{Collector}&Interstate & Interstate& Interstate\\
         &&Dry&\textcolor{red}{Wet}&\textcolor{red}{Snowy} & Dry& Dry\\
         &&Non-intersection&Non-intersection&Non-intersection & Non-intersection& Non-intersection\\
         &&Non-work zone&\textcolor{red}{Work zone}&\textcolor{red}{Work zone} & Non-work zone& Non-work zone\\
         &&Overhead bridges&\textcolor{red}{Open roads}&Overhead bridges & Overhead bridges& Overhead bridges\\\hline

         \multirow{7}{*}{f}&\multirow{7}{*}{\includegraphics[width=5cm, height=2cm, keepaspectratio]{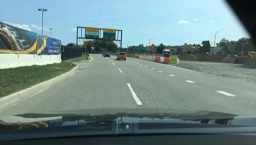}}&Daytime&\textcolor{red}{Dawn}&\textcolor{red}{Dawn}&Daytime&Daytime\\
         &&Clear&\textcolor{red}{Snowing}&Clear & Clear& Clear\\ 
         &&Interstate&\textcolor{red}{Collector}&\textcolor{red}{Local road} & \textcolor{red}{Arterial}& \textcolor{red}{Arterial}\\
         &&Dry&\textcolor{red}{Snowy}&Dry & Dry& Dry\\
         &&Non-intersection&Non-intersection&\textcolor{red}{4-way} & Non-intersection& Non-intersection\\
         &&Work zone&Work zone&\textcolor{red}{Non-work zone} & Work zone& Work zone\\
         &&Open roads&Open roads&Open roads & Open roads& Open roads\\\hline

         \multirow{7}{*}{g}&\multirow{7}{*}{\includegraphics[width=5cm, height=2cm, keepaspectratio]{rain.png}}&Daytime&\textcolor{red}{Dawn}&Daytime&\textcolor{red}{Dawn}&Daytime\\
         &&Rainy&\textcolor{red}{Clear}&Rainy & Rainy& Rainy\\ 
         &&Local road&\textcolor{red}{Collector}&Local road & Local road& Local road\\
         &&Wet&\textcolor{red}{Snowy}&Wet & Wet& Wet\\
         &&Non-intersection&Non-intersection&\textcolor{red}{4-way} & Non-intersection& Non-intersection\\
         &&Non-work zone&\textcolor{red}{Work zone}&Non-work zone &Non-work zone&Non-work zone\\
         &&Open roads&Open roads&Open roads & Open roads& Open roads\\\hline
    \end{tabular}
\end{table*}

To supplement the ablation study in the preceding section, seven examples (a-g) are further provided, with their corresponding results detailed in TABLE \ref{tab:ablat2}. The baseline model demonstrates insufficient knowledge about driving scene identification. It fails in multi-label scene classification,  only correctly identifying each scene with at most three attributes. 

The foundation model, which assimilates knowledge regarding driving scene identification from diverse single-label datasets, significantly outperforms the baseline model that lacks such knowledge, correctly identifying scenes with respect to four to six attributes. However, an observable limitation of the foundation model is its frequent misidentification of dry road surface conditions as snowy, daytime as dawn, non-work zones as work zones, other road types as local roads, and non-intersection or three-way intersections as four-way intersections. For instance, the foggy weather condition in scenario (d) may have led the foundation model to erroneously classify daytime as dawn and the dry road surface as snowy. Furthermore, the vehicle's close proximity to the intersection makes it difficult to differentiate three-leg and four-leg features. Scenario (f) depicts an interstate highway that is slightly more complex than typical interstate driving scenes. The advertisement banner on the left and the open construction space on the right likely lead the model to misidentify it as a four-way intersection on a local road. Additionally, the atypical work zone features and the relatively farther distance of the safety drums from the ego-vehicle fail the model in recognizing the work zone scenario. Overall, certain combinations of scene attribute labels are less prevalent in single-labeled datasets, thereby limiting the foundation model's exposure and subsequent ability to accurately identify these specific scenes.

Utilizing feature proximity to recommend new training samples that are similar to the test data, CAL adapts the foundation model quickly to multi-attribute sence identification, demonstrating superior ability in classifying driving scenes comprehensively, as illustrated in the two rightmost columns in TABLE \ref{tab:ablat2}. Notably, the model adapted via CAL with only two iterations corrects most misclassifications by the foundation model, demonstrating the effectiveness of CAL. Meanwhile, with iterations up to five, the classification model learns the more fine-grained feature distributions, thus improving its ability to identify driving scenes. For instance, in scenes a, b, d, and e, the foundation model misclassified the weather-related road condition as snowy. Assisted by CAL, now the adapted model can correctly recognize the category and reassign the class label. Furthermore, CAL mitigates dominant biases in feature distributions across single-label classification datasets. Specifically, in recognizing a four-way intersection, the model relies more on the complex behavior of traffic participants rather than on the actual geometric layout of the intersection (as seen in scenes c and g). CAL addresses this by providing additional class labels to supervise the learning of features that were previously either unseen or unaddressed in monotask learning.

\subsection{Comparison to SOTA Multitask Classification Models} 
\label{subsection:Comparison to SOTA Multitask Classification Models}

In the end, the proposed KAA-CAL learning method was compared to SOTA models reviewed in Section \ref{subsec:Multi-label Driving Scene Classification}. To demonstrate the generalization of the KAA-CAL learning method, the comparative study was conducted on both BDD100K \cite{yu2020bdd100k} and HSD \cite{narayanan2019dynamic}. BDD100K has labels for three driving scene attributes, including Time of Day, Weather, and Scene Type. The HSD dataset includes labels for four attributes: Intersection Type, Surface Condition, Driving Types, and Weather. Despite the limited number of attributes provided, these two datasets offer an opportunity for objectively evaluating KAA-CAL against SOTA methods. TABLE \ref{tab:DATASET} summarizes the sizes of these datasets in terms of training, validation, and testing partitions.  

\begin{table}[!ht]
    \centering
    \caption{BDD100K and HSD Datasets}
        \begin{tabular}{l|>{\centering\arraybackslash}p{0.6in}>{\centering\arraybackslash}p{0.6in}>{\centering\arraybackslash}p{0.5in}}
            %\hline
            \toprule
            Dataset& Training & Validation & Testing \\
            \midrule
            %\hline
            BDD100K & 70,000 & 7,000 & 3,000  \\
            HSD$^\ast$ &53,115 & 5,500 & 2,275  \\
            \bottomrule
            \multicolumn{4}{l}{$^\ast$Scene images were sampled every 100 frames from 80 hours}\\
            \multicolumn{4}{l}{$\;$ of driving videos.}
        \end{tabular}
        \label{tab:DATASET}
\end{table}

Our multi-attribute scene identification model utilizes the foundation model as its deep encoder. It was subsequently refined via CAL to identify driving scenes in BDD100K and HSD, respectively, using a total of 15\% of their training data over three iterations. In contrast, the four SOTA models, whose encoders were pretrained on ImageNet, were trained on the full BDD100K and HSD training datasets, respectively. Results of the comparative study are summarized in TABLE \ref{tab:BDD and Honda}.

\begin{table*}[htbp]
    \centering
    \caption{Classification accuracy (\%) comparison with SOTA methods on BDD100K and Honda HSD dataset}
    \begin{tabular}{>{\raggedright\arraybackslash}p{1in}|>{\centering\arraybackslash}p{0.6in}>{\centering\arraybackslash}p{0.6in}>{\centering\arraybackslash}p{0.6in}|>{\centering\arraybackslash}p{0.6in}>{\centering\arraybackslash}p{0.85in}>{\centering\arraybackslash}p{0.8in}>{\centering\arraybackslash}p{0.6in}}
    \toprule
       \multirow{2}{*}{}&\multicolumn{3}{c|}{BDD100K} & \multicolumn{4}{c}{HSD}\\
       \cmidrule{2-8}
       %\cline{2-8}
      
        & Time of Day & Weather & Scene Type  &  Intersection & Surface Condition & Driving Type & Weather \\
        \midrule
        ResNet18 \cite{MUlti5} & \underline{92.7} & 80.6 & 77.7&  \underline{88.9} & \underline{91.0} & 80.5 & 67.1 \\
        ViT-B-16 \cite{MUlti5}& 92.2 & 80.1 & 76.9&  88.0 & \underline{91.0} & \underline{81.1} & \underline{68.3} \\
        CF-NET \cite{multi2}& \underline{92.7} & \underline{81.0} & \underline{78.5}&  \textbf{89.0} & 90.8 & 78.2 & 66.7 \\
        RECNet \cite{RECNET}& 92.6 & 80.2 & 76.7&  87.0 & 90.4 & 80.7 & 68.0 \\
        KAA-CAL (ours)$^\ast$ & \textbf{92.7} & \textbf{81.8} & \textbf{78.6}&  87.8 & \textbf{91.1} & \textbf{81.4} & \textbf{68.6}  \\
        \bottomrule
    \end{tabular}
    \raggedright\\
    The best performance is bolded, and the second best is underlined. $\ast$ trained using only 15\% of the training data. 
    \label{tab:BDD and Honda}
\end{table*}

KAA-CAL, which uses only 15\% of the training data, achieves task accuracy comparable to SOTA methods that utilize all the training data. Specifically, on BDD100K, KAA-CAL achieves the same highest accuracy as ResNet-18 and CF-NET in identifying Time of Day. KAA-CAL's accuracy in identifying Weather (81.8\%) exceeds the second-best method, CF-NET (81.0\%), by 0.8\%, and it slightly surpasses CF-NET by 0.1\% in identifying Scene Type. 

On HSD, KAA-CAL achieves the highest accuracy in identifying Surface Condition (91.1\%), Driving Type (81.4\%), and Weather (68.6\%), exceeding the second best method by a small margin ranging from 0.1\% to 0.3\%. However, its accuracy in identifying Intersection Type (87.8\%) is 1.2\% below the best method, CF-NET (89.0\%). 

Notably, Scene Type in BDD 100K and Driving Type in HSD are not attributes that the foundation learned to recognize. With the foundation model's comprehensive knowledge about driving scenes, and facilitated by CAL, the model efficiently gains the knowledge to identify driving scenes using these new attributes. While SOTA models listed in TABLE \ref{tab:BDD and Honda} attain the performance comparable to KAA-CAL, they are not ready to characterize driving scenes using other attributes.

\section{Conclusion}\label{conclusion}

This paper presents KAA-CAL, a deep learning method for multi-attribute driving scene identification, a fundamental yet challenging visual perception capability for AVs. KAA improved the average classification accuracy by 31.3\% compared to the baseline model pretrained on ImageNet. CAL further boosts the performance to 92.7\%. Additionally, KAA-CAL outperforms SOTA methods on BDD100K and HSD,  achieving this while using only 15\% of the training data and even recognizing attributes for which the foundation models was not previously train. 

KAA-CAL lays a methodological foundation for not only multi-attribute scene identification - a high level perception - but other vision capabilities that AV needs. A universal vision foundation model can be built by synergizing the knowledge of those constituents. Due to a deliberate focus on the methodological development, KAA-CAL was not evaluated on a broader range of datasets. Yet, the number of scene attributes can be scaled up in this foundation model by following the data preparation and learning approach outlined in this paper. Furthermore, despite its efficiency, CAL has not yet achieved the ideal few-shot model refinement for downstream tasks. Future studies can also leverage the domain-specific prior and semantic guidance embedded in the driving scene identification model with vision-language models to advance video understanding and reasoning for AVs. Nevertheless, the work presented in this paper lays the groundwork for exploring these opportunities for generalization, scaling, improvement, and extension.  

\section*{Acknowledgment}
Qin receives funding from the Rural Safe Efficient Advanced Transportation Center, a Tier-1 University Transportation Center funded by the United States Department of Transportation (USDOT), through agreement number 69A3552348321. The contents of this paper reflect the views of the authors. USDOT assumes no liability for the contents or use thereof.
 
\bibliographystyle{IEEEtran} 
\bibliography{IEEEexample}

%\newpage

%\vspace{-33pt}
\vspace{-30pt}  
\begin{IEEEbiography}
[{\includegraphics[width=1in,height=1.25in,clip,keepaspectratio]{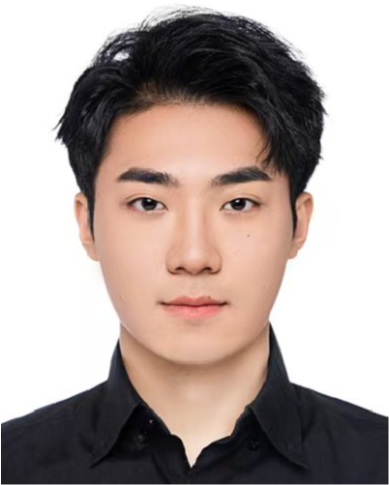}}]{Ke Li} (Graduate Student Member, IEEE) received the B.S. degree in transportation from Southeast University, China, in 2021, and the M.S. degree in Civil Engineering from University of Illinois Urbana-Champaign, IL, USA, in 2023. He is currently pursuing the Ph.D. degree in the Department of Civil Engineering at Stony Brook University, NY, USA. He has been a research assistant since 2023, working on perception and recognition for autonomous driving systems and intelligent systems using deep learning methods.
\end{IEEEbiography}
\vspace{-30pt}  
\begin{IEEEbiography}
[{\includegraphics[width=1in,height=1.25in,clip,keepaspectratio]{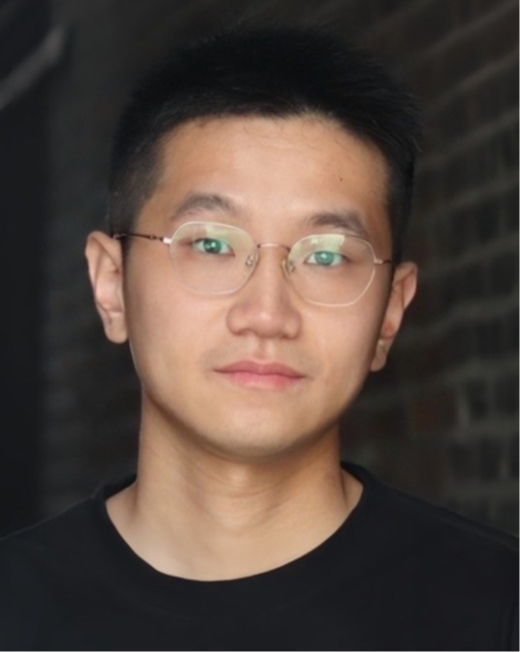}}]{Chenyu Zhang} (Graduate Student Member, IEEE) received the B.S. degree in Bridge Engineering in 2018 and the M.S. degree in Civil Engineering in 2021, both from the School of Highway, Chang’an University, Xi’an, China. He is currently pursuing the Ph.D. degree in Civil Engineering at Stony Brook University, Stony Brook, NY, USA. His research interests include intelligent transportation systems, transportation asset management, and resilient infrastructure systems.
\end{IEEEbiography}
\vspace{-30pt}  
\begin{IEEEbiography}
[{\includegraphics[width=1in,height=1.25in,clip,keepaspectratio]{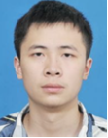}}]{Yuxin Ding}

received his Bachelor degree from the Civil Aviation University Of China, and Master degree from Georgia Institute Of Technology. He is currently pursuing the Ph.D. with the Department of Civil and Environmental Engineering, Penn State University, Pennsylvania, USA. His research is focused on autonomous vehicles and dynamic system modeling. 

\end{IEEEbiography}
\vspace{-30pt}  
\begin{IEEEbiography}[{\includegraphics[width=1in,height=1.25in,clip,keepaspectratio]{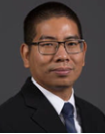}}]{Xianbiao Hu}
received the B.Eng. and M.Eng. degrees in Transportation Engineering from Tongji University, China, and the Ph.D. degree in Transportation Engineering from the University of Arizona, USA. His research focuses in the area of Smart Mobility System, Dynamic System Modeling, Vehicle Technology, Active Demand Management, Automated Vehicles, and Transportation Electrification.
 \end{IEEEbiography}
\vspace{-30pt}  

\begin{IEEEbiography}
[{\includegraphics[width=1in,height=1.25in,clip,keepaspectratio]{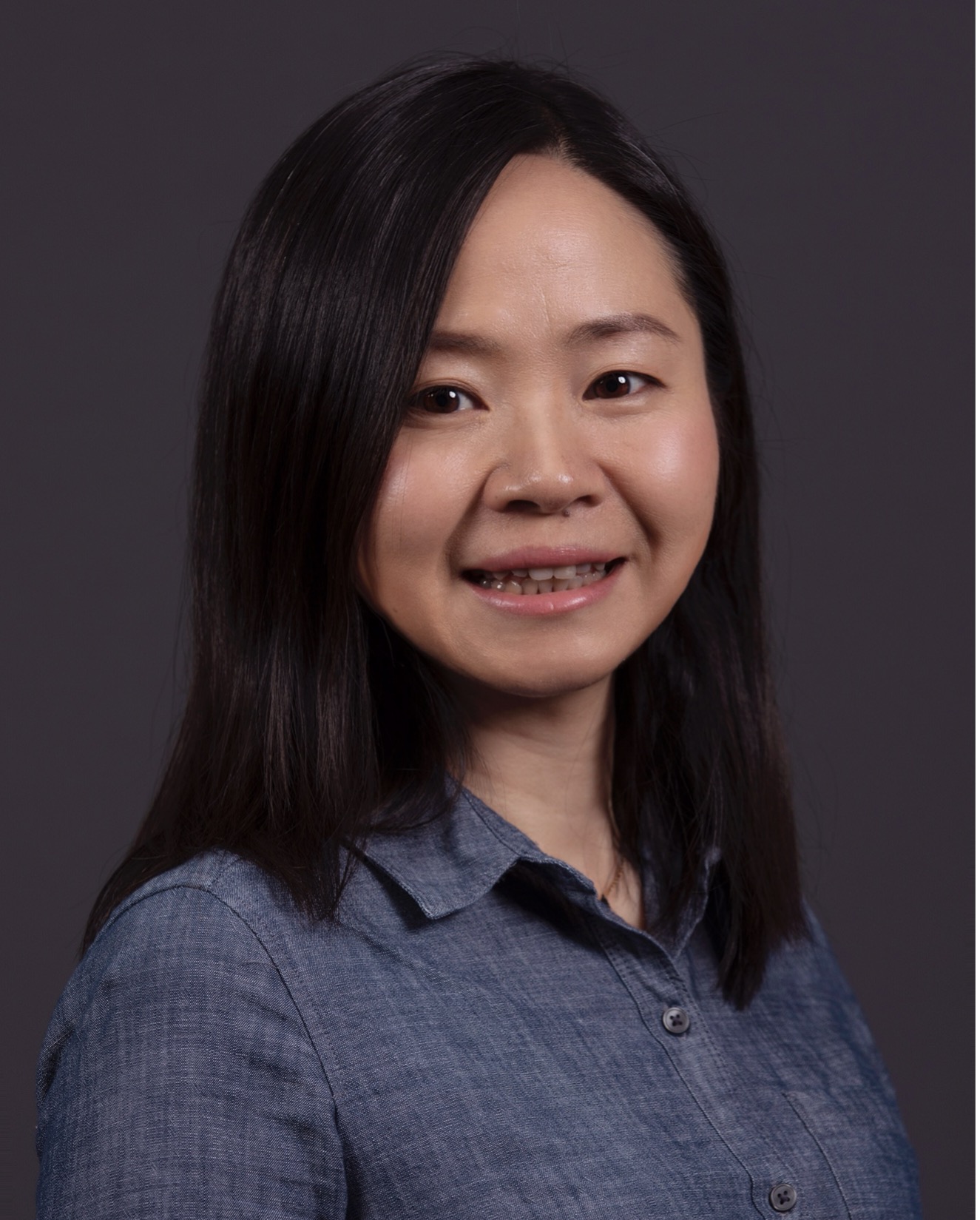}}]
{Ruwen Qin} (Member, IEEE) received the B.E. and M.S. degrees in spacecraft design from Beijing University of Aeronautics and Astronautics and the Ph.D. degree in industrial engineering \& operations research from Pennsylvania State University. She is an Associate Professor of civil engineering at Stony Brook University. Her research is focused on sensing and deep learning methods to build perception and cognition abilities for intelligent systems like autonomous vehicles. She is a member of IEEE Intelligent Transportation Systems Society.      
\end{IEEEbiography}

\vfill

\end{document}